%% file: main.tex
\documentclass[11pt]{article}

\usepackage[final]{acl}

\usepackage{times}
\usepackage{latexsym}
\usepackage{caption}
\usepackage{hyperref}
\usepackage{multirow}
\usepackage{makecell}
\usepackage{color}
\usepackage{xcolor}
\usepackage{colortbl}
\usepackage{bm}
\usepackage{tcolorbox}
\usepackage{wrapfig}
\usepackage{enumitem}
\usepackage{booktabs}
\usepackage{multirow}
\usepackage{amsmath}
\usepackage{amssymb}
\usepackage{algorithm}
\usepackage{algpseudocode} 
\usepackage{makecell}
\usepackage{pifont}
\usepackage{CJK}

\usepackage[T1]{fontenc}

\usepackage[utf8]{inputenc}

\usepackage{microtype}

\usepackage{inconsolata}

\usepackage{graphicx}

\newcommand\blfootnote[1]{%
  \begingroup
  \renewcommand\thefootnote{}\footnote{#1}%
  \addtocounter{footnote}{-1}%
  \endgroup
}

\title{TEMA: Anchor the Image, Follow the Text for Multi-Modification Composed Image Retrieval}

\author{Zixu Li$^{1}$\quad Yupeng Hu$^{1*}$\blfootnote{~corresponding authors}\quad Zhiheng Fu$^{1}$\quad Zhiwei Chen$^{1}$\quad Yongqi Li$^2$\quad Liqiang Nie$^{3}$ \\
$^1$ School of Software, Shandong University \quad \\
$^2$ Department of Computing, Hong Kong Polytechnic University \\
$^3$ School of Computer Science and Technology, Harbin Institute of Technology (Shenzhen) \quad
\\
{\tt\small \{lizixu.cs, fuzhiheng8, zivczw, liyongqi0, nieliqiang\}@gmail.com \quad huyupeng@sdu.edu.cn} \\ 
}

\begin{document}
\maketitle
\input{2-sec/0_abstract}

\blfootnote{$^*$ Corresponding Author: Yupeng Hu}

\input{2-sec/1_intro}

\input{2-sec/2_related}
\input{2-sec/3_method}
\input{2-sec/4_exp}
\input{2-sec/5_conclusion}
\input{2-sec/6_limitations}
\input{2-sec/7_Ethical}

\input{2-sec/supplementary}

\clearpage

\bibliography{arr}

\end{document}

%% file: 2-sec/0_abstract.tex
\begin{abstract}
Composed Image Retrieval (CIR) is an important image retrieval paradigm that enables users to retrieve a target image using a multimodal query that consists of a reference image and modification text. Although research on CIR has made significant progress, prevailing setups still rely simple modification texts that typically cover only a limited range of salient changes, which induces two limitations highly relevant to practical applications, namely \textbf{Insufficient Entity Coverage} and \textbf{Clause-Entity Misalignment}. In order to address these issues and bring CIR closer to real-world use, we construct two instruction-rich multi-modification datasets, \textbf{M-FashionIQ} and \textbf{M-CIRR}. In addition, we propose \textbf{TEMA}, the Text-oriented Entity Mapping Architecture, which is the first CIR framework designed for multi-modification while also accommodating simple modifications. Extensive experiments on four benchmark datasets demonstrate that TEMA's superiority in both original and multi-modification scenarios, while maintaining an optimal balance between retrieval accuracy and computational efficiency. Our codes and constructed multi-modification dataset (\textbf{M-FashionIQ and M-CIRR}) are available at \href{https://github.com/lee-zixu/ACL26-TEMA/}{https://github.com/lee-zixu/ACL26-TEMA/}
\end{abstract}

%% file: 2-sec/1_intro.tex
\section{Introduction}
\label{sec:intro}

\begin{figure}[t!]
  \begin{center}
\includegraphics[width=\linewidth]{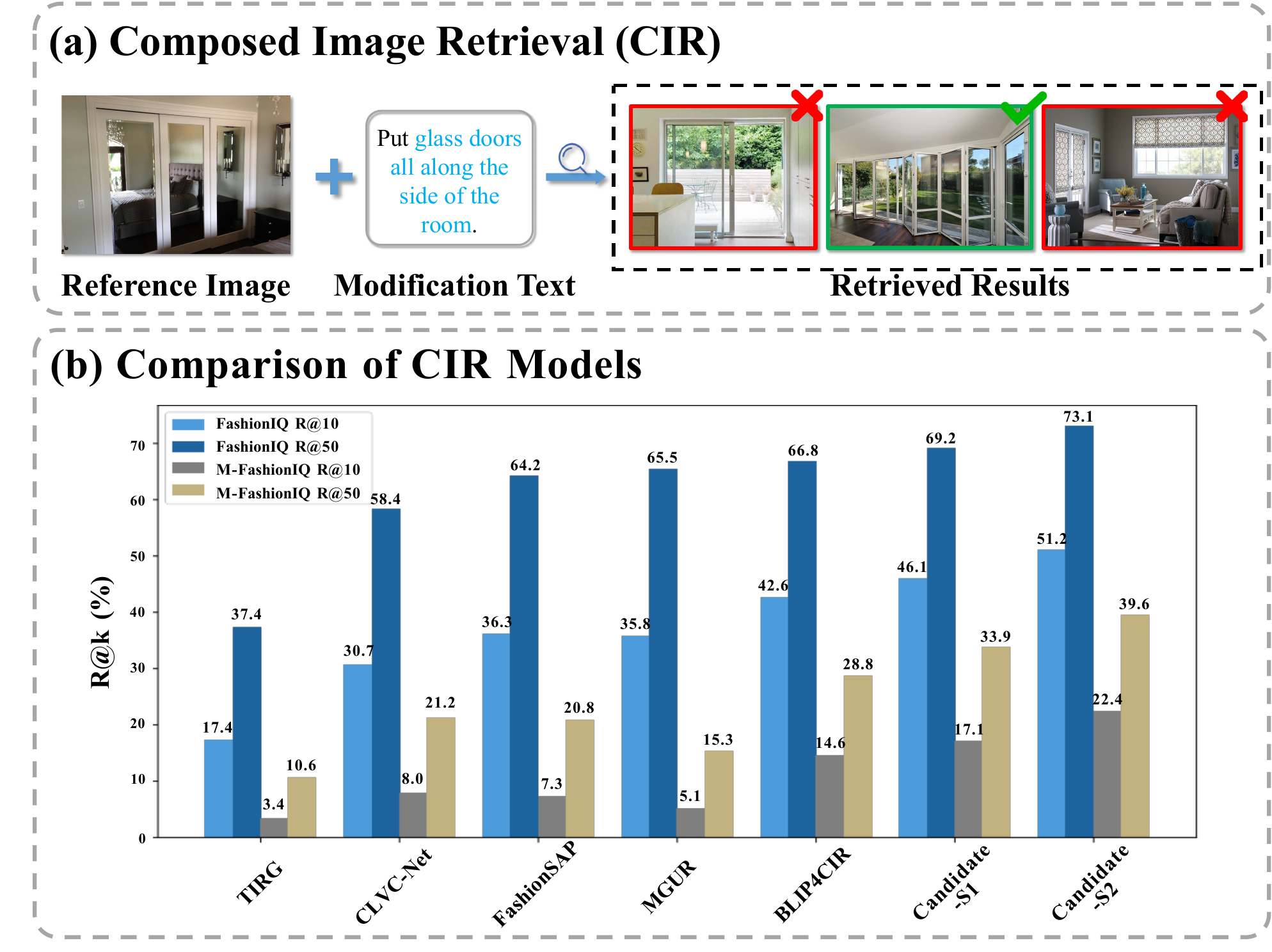}
  \end{center}
\vspace{-12pt}
  \caption{(a) Example of traditional CIR, and (b) Performance comparison of representative baselines on CIR datasets in original and multi-modification scenarios (all models are trained on original FashionIQ).
  }
\vspace{-8pt}
  \label{fig:intro-1}
\end{figure}

Composed Image Retrieval (CIR)~\cite{INTENT,HINT,PAIR,OFFSET,MEDIAN} uses a ``reference image $+$ modification text'' query to locate target images that satisfy the user's retrieval intent within large image collections. As shown in Figure~\ref{fig:intro-1}(a), unlike text only retrieval, CIR leverages the reference image to provide visual priors such as appearance, layout, and style, while the modification text specifies how to modify it relative to this reference anchor. CIR models often need to preserve the subject and style while imposing multiple attribute and relation constraints on multiple entities in order to achieve precise retrieval~\cite{liu2021distilling,xie2026delving,liu2021quadrupletbert,HUD,REFINE}. This paradigm has substantial application value in multimodal learing~\cite{xiao2025visual,zheng2025mma,song2,liu2024synthvlm,song13}, human-computer interaction~\cite{Long_2026,zhou2024boosting,li2026whatsmissingscreentoactionuiintheloop,zhou2022understanding}, and it has attracted broad attention in recent years~\cite{acl-1}.

In recent years, although research on CIR has made notable progress, prevailing setups~\cite{ENCODER,HABIT,INTENT,MELT} still rely on short modification texts that typically cover only a small number of salient changes ~\cite{FineCIR, cola}. This reliance gives rise to two limitations that are highly relevant to practical applications. \textbf{(1) Insufficient Entity Coverage}. When multiple to-be-modified entities are present, the training signal tends to concentrate on salient regions and omit some entities. In the modification texts used for CIR, detailed descriptions account for more than $80\%$ on average, and additional portions are occupied by prepositions and conjunctions. The proportion explicitly referring to to-be-modified entities is small and can be easily ignored by models. \textbf{(2) Clause-Entity Misalignment}. In real applications, CIR is often used in image retrieval scenarios with stringent requirements for fine-grained details, whereas scenarios with lower requirements can be handled by unimodal image retrieval. It is therefore common for multiple modification clauses to constrain the same entity (e.g., simultaneously modifying the hem, shoulder embellishment, and belt of a dress), or for a single modification clause to constrain multiple entities of the same type (e.g., changing three retrievers in the image into huskies). 

However, we regret to observe that existing CIR models struggle to meet multi-modification requirements in practical settings. As shown in Figure~\ref{fig:intro-1}(b), we convert samples from the FashionIQ validation set into a multi-modification form and evaluate several strong CIR baselines~\cite{tirg, clvcnet, fashionsap, mgur, blip4cir, candidate,TME}. We find a pronounced performance drop under multi-modification scenarios, which is likely due to the lack of multi-modification annotations during training and a heightened susceptibility to the limitations of Insufficient Entity Coverage and Clause-Entity Misalignment. To address these issues and realize the two core capabilities of entity coverage and multi-clause aggregation, thereby advancing CIR toward real-world applications, we propose a complementary \textbf{\textcolor[HTML]{A32D26}{data}} and \textbf{\textcolor[HTML]{2C77B9}{model}} solution.

\textbf{\textcolor[HTML]{A32D26}{Fresh Data Annotation}}: Without altering the original reference and target images or evaluation protocols, we expand the modification texts in FashionIQ~\cite{FashionIQ} and CIRR~\cite{cirr} into instruction-intensive multi-modification versions, constructing the \textbf{M-FashionIQ} and \textbf{M-CIRR} datasets. The new data replaces the original short texts with \textbf{Multi-Modification Texts (MMT)}, generated by MLLM and verified by human annotators, explicitly presenting constraint structures with multiple entities and clauses. This approach provides more comprehensive entity clues and denser training signals for the ``Insufficient Entity Coverage'' challenge, while offering a test environment more aligned with practical applications for the ``Clause-Entity Misalignment'' challenge. The aim is to create benchmarks that are closer to real-world scenarios, rather than simply improving performance. The multi-modification annotations, though increasing the complexity of understanding, are more aligned with practical applications and contribute to the real-world deployment of CIR.

\textbf{\textcolor[HTML]{2C77B9}{Novel Model Architecture}}: We propose the first CIR framework for multi-modifications while accommodating simple modifications, named \textbf{TEMA} (\textbf{T}ext-oriented \textbf{E}ntity \textbf{M}apping \textbf{A}rchitecture). To address the "Insufficient Entity Coverage" problem, we design the \textit{MMT Parsing Assistant (PA)}, which enhances the exposure and coverage of modified entities during training through summarization and consistency checks. During inference, the PA is disabled to avoid additional dependencies and delays. To tackle the ``Clause-Entity Misalignment'' issue, we design an \textit{MMT-oriented Entity Mapping module (EM)} that introduces learnable queries, consolidating multiple clauses of the same entity on the text side and aligning them with the corresponding visual entities on the image side. This stabilizes the modeling of ``one-to-many'' relationships without explicit alignment annotations. The collaboration of these two modules enables the model to acquire transferable entity coverage and aggregation abilities while remaining robust to multi-granularity multimodal query instructions.

The main contributions are as follows:

\noindent
$\bullet~$We find that existing CIR models struggle with the multi-modification requirements in practical scenarios. To address this, we construct two instruction-intensive multi-modification datasets, M-FashionIQ and M-CIRR.

\noindent
$\bullet~$We propose the first CIR framework that accommodates both original and multi-modification scenarios, TEMA, which can learn transferable entity coverage and aggregation abilities during training while maintaining robustness for multi-granularity multimodal query instructions.

\noindent
$\bullet~$Our proposed TEMA achieves optimal performance in both original (FashionIQ and CIRR datasets) and multi-modification (M-FashionIQ and M-CIRR datasets) CIR scenarios. A large number of quantitative and qualitative experiments validate its superiority.

%% file: 2-sec/2_related.tex
\section{Related Work}

\noindent\textbf{Composed Image Retrieval.}
This task aims to retrieve target images based on a reference image and modification text. Existing CIR methods can be broadly categorized into traditional models~\cite{tirg, mgur, val, cosmo, clvcnet} and VLP-based models~\cite{lfclip, clip4cir-v2, limn, fashionern, ssn}. Recently, the rapid advancement of Large Vision-Language Models (LVLMs)~\cite{he2024robust,sun2023hierarchical,pu2025she,pu2025robust} and visual foundation models~\cite{tan2026blend,jiang2026foeforesterrorsmakes,song5,zheng2025adamcot,li2025faithact} has dramatically enhanced cross-modal understanding~\cite{liu2025fusion,lin2026mmfinereason,li2025taco,zhong2026collaborativemultiagentscriptsgeneration} and instruction-following capabilities~\cite{liu2025uniform,xiao2026not,yang2026infactdiagnosticbenchmarkinduced,liu2026chartverse,xiao2025prompt}. However, despite the powerful representation abilities brought by these advancements, existing CIR frameworks are mostly limited to addressing simple modification requests. To bridge this gap, our proposed multi-modification datasets facilitate more comprehensive modification descriptions through MMT, thereby better satisfying users' detailed, instruction-driven retrieval intentions in practical application scenarios.

\noindent\textbf{Multi-object and fine-grained annotations}.
As user retrieval needs become more complex~\cite{tian2025core,huangnodes,huang2023robust,tian2025open,xu2025hdnet,lu2024robust,zhou2025dragflow,lu2025does,huang2025enhancing,liu2024dual,sun2023stepwise}, modification text annotations must evolve to support multi-object and fine-grained descriptions. Driven by the strong semantic parsing and reasoning capabilities of modern Large Language Models (LLMs)~\cite{chang2026balora,an2025amo,yuan2026strucsum,huang2024exploring,song8}, exploring complex, multi-granular textual modifications has become a new trend~\cite{wang2025stability,jiang2025self,li2025curriculum,ERASE}. While several CIR studies have made progress in this direction, common limitations persist. Works like Cola~\cite{cola}, MagicLens~\cite{magiclens}, and ReT-2~\cite{ret-2} primarily examine multi-object interference, whereas MIST~\cite{mist} and early CTI-IR~\cite{JointAttribute} construct training data without considering multi-modification requirements. Even methods like FineCIR~\cite{FineCIR}, which explicitly parses modification semantics, fail to guarantee that the modification text covers all to-be-modified entities. In contrast, our TEMA explicitly targets multi-modification CIR by introducing MMT together with PA and EM, effectively bridging the gap in explicitly modeling multi-entity to multi-clause alignment.

%% file: 2-sec/3_method.tex
\section{Multi-Modification CIR Datasets Construction}
\label{sec:benchmark}
In this section, we introduce the constructed multi-modification dataset. Note that our goal is to create benchmarks that are closer to real-world scenarios rather than simply improving performance. 

\begin{figure*}[ht!]
\centering
\includegraphics[width=0.95\linewidth]{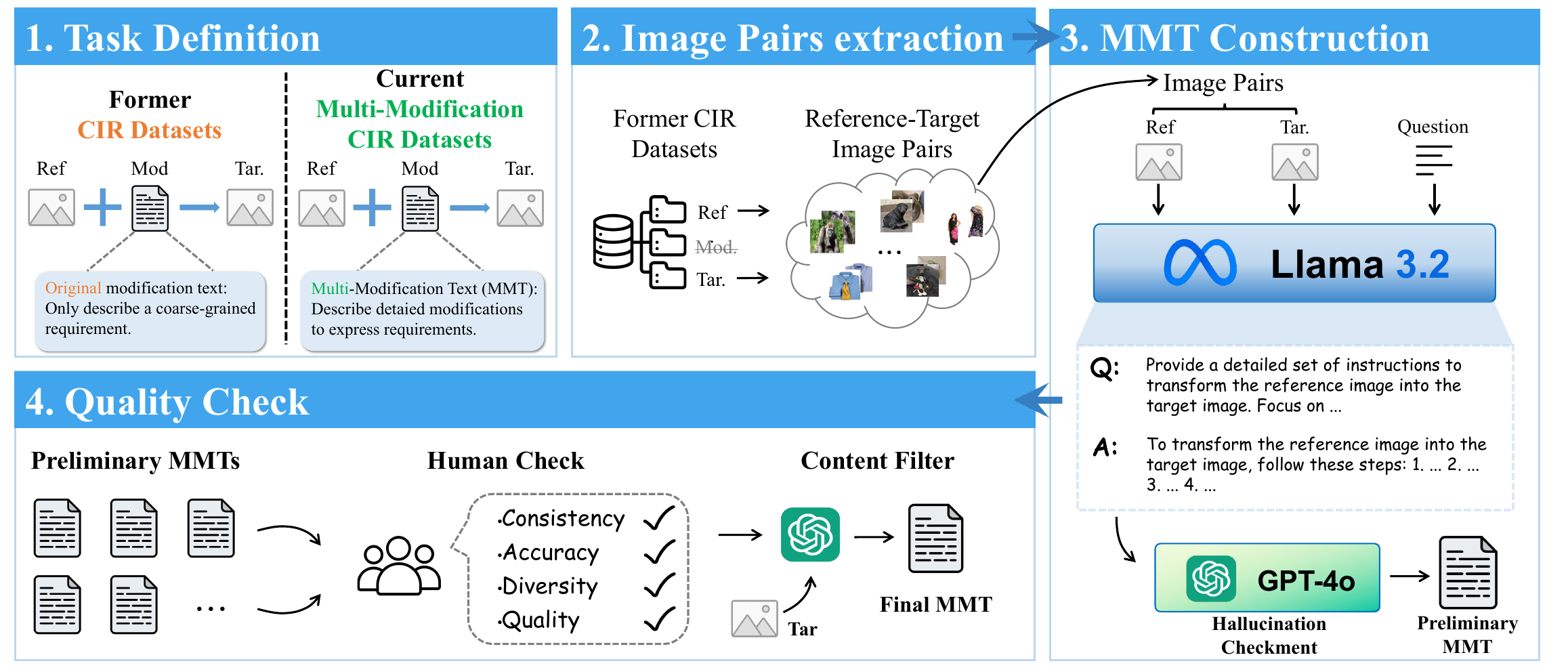}
  \vspace{-10pt}
  \caption{Pipeline of the construction of our proposed multi-modification CIR datasets.}
  \vspace{-17pt}
  \label{fig:generate}
\end{figure*}

To promote CIR tasks closer to practical application, we construct two datasets: a fashion-domain dataset M-FashionIQ, and an open-domain dataset M-CIRR. 
They are built upon the classic CIR datasets FashionIQ and CIRR. 
Leveraging the automatic annotations generated by MLLM, we incorporate a manual review process to ensure high quality of the datasets.
Our empirical results demonstrate that this combined approach effectively captures more nuanced users' modification requests while minimizing false-negative samples, thus enhancing the datasets' suitability for training and testing in multi-modification scenarios.

\noindent \textbf{Data Construction.}
\label{sec:data_cons}
Since the primary distinction between multi-modification datasets and original CIR datasets lies in the modification text, we select two classical CIR datasets (FashionIQ and CIRR) and re-label the modification text in the original triplets. 
We note the powerful multimodal comprehension capabilities~\cite{huangrevisiting,huang2025final,song6,ma2026stableexplainablepersonalitytrait,song7} of Multimodal Large Language Models (MLLMs)~\cite{llama-3}, and therefore we utilize an MLLM, Llama 3.2~\cite{llama-3} as our primary automatic annotation tool.
Specifically, as illustrated in Figure~\ref{fig:generate}~(2) and (3), we extract triplet samples from the original datasets and utilize the reference-target image pairs as the input to Llama 3.2.
Simultaneously, we design detailed prompts that necessitate the MMT generated by the Llama 3.2 to faithfully adhere to the original modification texts while articulating refined modification requests that specify the intricacies of transforming the reference image to the target image, as shown in \textbf{Q} in Figure~\ref{fig:generate}~(3). 
This requires the MLLM to understand both the reference and target images, and describe their differences, outputting the candidate MMT.

Moreover, since the two datasets belong to diverse domains, we also design prompts tailored to the specific characteristics of each dataset. For the FashionIQ, we require the MMT generated by Llama 3.2 to focus on various aspects of clothing (\textit{e.g.}, shape, color). In contrast, for the CIRR, we emphasize the different objects present within the open scenario. 
Such tailored prompts maximize attention on the unique dataset characteristics, ensuring that the generated MMT closely aligns with the authentic modification texts encountered in real retrieval scenarios~\cite{li2024optimizing,zhou2024adversarial,hu2023semantic,xie2026conquer}. Following the above process, we obtain the automatically generated MMT. Besides, we provide the detailed prompts used for both datasets, along with comparative analysis of the impacts on the MMT generated from various prompts (detailed in \textbf{Appendix}~\ref{sup:prompt}).

After obtaining the MMT, we further refine the text output. 
Considering the hallucination issues~\cite{hallu_survey1} in current MLLMs, we specifically aim to eliminate hallucinated content embedded within the MMT by the Llama 3.2.
Specifically, we use GPT-4o~\cite{gpt} (note that other large language models such as Llama-3~\cite{llama-3} can also achieve similar results) to perform a hallucination check on the previously obtained MMT. 
This process detects and removes any obvious hallucinated content in the text, resulting in the preliminary
MMT that can be further processed for the quality check process.

\noindent \textbf{Quality Check.}
After obtaining the Preliminary MMT, we further adopt a hybrid quality check process involving both human and machine efforts~\cite{lin2025se,song3,wang2026fbs,wang2026tracking} to ensure the overall quality of the MMTs. Specifically, to reduce the workload of human annotators, we first conduct a manual review solely based on textual content, without referencing the associated images. In this stage, a team of 10 research assistants is instructed to examine and revise the texts from four perspectives, including \textit{Consistency}, \textit{Accuracy}, \textit{Diversity}, and \textit{Quality}, which are detailed in \textbf{Appendix}~\ref{sup:quality_check}.
While ensuring the linguistic quality of each MMT, it is equally important to verify whether the modification is faithful to the corresponding reference image. To address this issue, we introduce a \textit{Content Filter} stage following the manual refinement (detailed in \textbf{Appendix}~\ref{sup:content_filter}).

\noindent \textbf{Positive and negative samples.}
Essentially, the positive samples in our multi-modification dataset are justifiable since we directly replace the original modification texts with MMT while retaining the original reference and target images in each triplet, and the generated MMT remains faithful to the original modification texts. Furthermore, as MMT provides more precise descriptions of the differences between the reference and target images while encompassing the original modification intent, our empirical evidence indicates that this extension method is effective and mitigates the issue of false negatives that originally existed in the CIR task datasets (detailed in \textbf{Appendix}~\ref{sup:false_negative}).

We present the dataset statistics in \textbf{Appendix}~\ref{sup:data_stat} and compare it with the original CIR dataset. And in \textbf{Appendix}~\ref{sup:metrics}, we introduce evaluation metrics.

\section{Method}
\label{sec:method}

\begin{figure*}[ht!]
  \begin{center}
  \includegraphics[width=\linewidth]{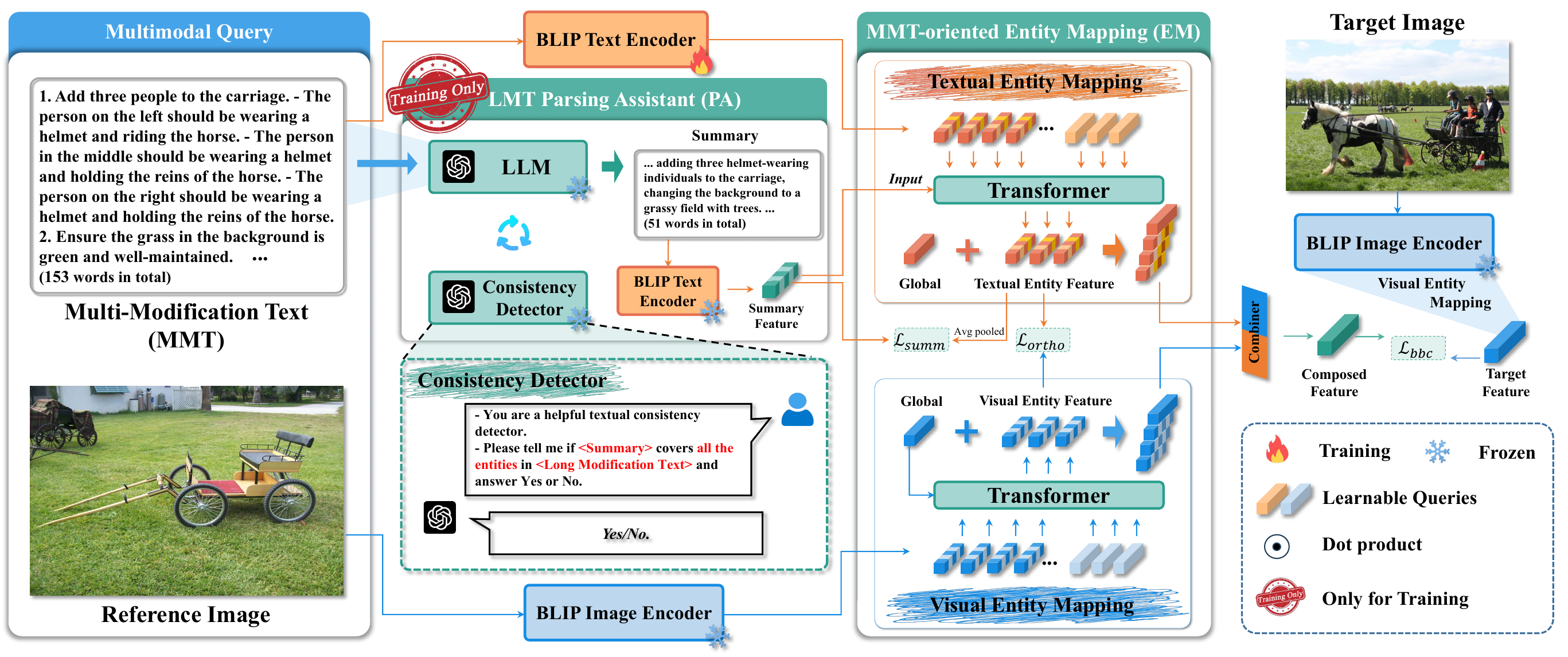}
  \end{center}
  \vspace{-15pt}
  \caption{Overall architecture of our proposed TEMA.}
    \vspace{-18pt}
  \label{fig:framework}
\end{figure*}

To tackle CIR with multi-modification, we propose a \textbf{T}ext-oriented \textbf{E}ntity \textbf{M}apping \textbf{A}rchitecture (\textbf{TEMA}), which focuses on understanding modification intentions in MMT, enhancing to-be-modified entity coverage, and exploring clause–entity alignment in multimodal queries to meet fine-grained retrieval needs. As shown in Figure~\ref{fig:framework}, TEMA comprises two main components: \textit{1) MMT Parsing Assistant (PA)}, which includes an \textit{LLM-based text summarizer} and \textit{a Consistency Detector} to extract to-be-modified entities from MMT and perform entity coverage checks to enhance feature exposure (\textbf{used only during training}, and detailed in Sec.~\ref{sec:pa});
\textit{2) MMT-oriented Entity Mapping (EM)}, which consists of \textit{Textual \& Visual Entity Mapping} to aggregate multiple MMT clauses related to the same entities, guided by the summary (detailed in Sec.~\ref{sec:em}). We begin with preliminaries in Sec.~\ref{sec:pre}, then we elaborate on TEMA's modules.

\subsection{Preliminaries}
\label{sec:pre}

Given a dataset $\mathcal{T}=\left\{\left(x_{r},t_{m}, x_{t}\right)_{n}\right\}_{n=1}^{N}$ of $N$ triplets, where each triplet consists of a reference image $x_r$, a corresponding MMT $t_m$, and a target image $x_t$, the CIR task aims to retrieve $x_t$ based on the composition of $x_r$ and $t_m$. The model is trained to learn a shared embedding space where the multimodal query $(x_r, t_m)$ is mapped close to its target $x_t$. Formally, this objective is $\mathcal{F}\left(x_r, t_m\right) \rightarrow \mathcal{F}\left(x_t\right)$, where $\mathcal{F}(\cdot)$ denotes the learned embedding function for both image and text. The training minimizes the distance between $\mathcal{F}(x_r, t_m)$ and $\mathcal{F}(x_t)$, while ensuring non-matching pairs are pushed apart in the embedding space.

\subsection{MMT Parsing Assistant (PA)}
\label{sec:pa}
Given that MMT contains extensive modification details with sparsely mentioned entities that may be ignored by the model, we propose the PA module to maintain entity focus. 
It comprises an LLM-based text summarizer for to-be-modified entity parsing and a \textit{Consistency Detector} for entity coverage checking, operating only during training.

\noindent \textbf{LLM-Based Text Summarizer.} 
Specifically, considering the exceptional text comprehension capabilities of LLMs, we leverage an LLM (gpt-3.5-turbo~\cite{gpt}) to generate MMT summaries. We use a simple prompt to include all the to-be-modified entities in the summary, as follows,

\begin{tcolorbox}
\small{
\texttt{\{{\color{blue} MMT}\} denotes the multi-modification text. Please summarize it into a sentence that covers all the to-be-modified entities.}}
\end{tcolorbox}

\noindent \textbf{Consistency Detector.} 
To address potential LLM hallucinations~\cite{nature, hallu_survey1}, we implement a \textit{Consistency Detector} that verifies the summary's entity coverage. Specifically, we use the LLM (gpt-3.5-turbo~\cite{gpt}) as a Consistency Detector (with a detailed prompt provided in \textbf{Appendix}~\ref{sup:prompt}) to check whether the summary includes all to-be-modified entities from the MMT, while ensuring no extraneous entities.
If inconsistencies are detected, the summary is iteratively refined until it passes verification, yielding the final summary $t_s$. The summary features are then extracted using a frozen BLIP text encoder $\varPhi_\mathbb{T}$, formulated as:
\begin{equation}
    \textbf{E}_s=\varPhi_\mathbb{T}\left(t_{s}\right).
\label{summ_fea}
\end{equation}

We show the quality of the summary generated by the PA module in Section~\ref{sec:pa_qua}.

\subsection{MMT-oriented Entity Mapping (EM)}
\label{sec:em}
Due to the numerous modification details in the MMT, a single to-be-modified entity may correspond to multiple modification clauses. To avoid clause–entity misalignment, we design the MMT-oriented Entity Mapping (EM) module based on PA. It extracts the one-to-many correspondence between entities and MMT clauses, integrating the modification requirements.
Specifically, EM incorporates Textual and Visual Entity Mapping components. The textual EM consolidates multiple MMT clauses corresponding to the same to-be-modified entity, guided by the summary.
Moreover, to ensure comprehensive entity information preservation in the text tokens generated by EM, we propose a summary-guided distillation strategy, which promotes the generated text tokens to closely align with the to-be-modified entities parsed by PA.

\noindent \textbf{Feature Extracting.}
Specifically, we first extract the features of the reference image and MMT. 
Due to the input token limits of CLIP text encoder, we use BLIP~\cite{blip}, which has been proven effective on CIR task~\cite{blip4cir, candidate}, to extract the global feature $\textbf{E}^{g}_r\!\in\! \mathbb{R}^{D}$ and local feature $\textbf{E}^{l}_r\!\in\! \mathbb{R}^{C\times D}$ of the reference image $x_r$, formulated as,
\begin{equation}
    \textbf{E}^{g}_r=\varPhi_\mathbb{I}^g\left(x_r\right), \textbf{E}^{l}_r=\operatorname{FC_{\mathbb{I}}}\left(\varPhi_\mathbb{I}^l\left(x_r\right)\right), 
\label{eq:summ}
\end{equation}
where $D$ is the hidden dimension. $\varPhi_\mathbb{I}^g$ and $\varPhi_\mathbb{I}^l$ are the last and penultimate layers of the BLIP image encoder, respectively. $\operatorname{FC_{\mathbb{I}}}$ is to align the hidden dimension of the local feature and global feature.
Similarly, we use BLIP to extract the global feature $\textbf{E}^{g}_m$ and local feature $\textbf{E}^{l}_m$ of MMT, and the global feature $\textbf{E}^{g}_t$ and local feature $\textbf{E}^{l}_t$ of target image.

\noindent \textbf{Textual \& Visual Entity Mapping.}
To extract the one-to-many correspondence between to-be-modified entities and MMT clauses, we introduce a set of learnable queries $\mathbf{a}_q = \{a_1, ..., a_k\}$, which, along with the summary feature $\textbf{E}_s$ (from Eqn~(\ref{eq:summ})) and MMT local features $\textbf{E}^{l}_m$, serve as inputs to the transformer model.
Since the summary feature includes all to-be-modified entities with minimal details, the learnable queries aggregate the corresponding MMT clauses for the same entity, guided by the summary, formulated as,
\begin{equation}
        \hat{\mathbf{a}}_q = \operatorname{Transformer}\left(\left[\textbf{E}_s, \textbf{E}^{l}_m, \mathbf{a}_q\right]\right),
        \label{eq:aq}
\end{equation}
where $\hat{\mathbf{a}}_q\!\!\in\!\mathbb{R}^{N\times D}$ denotes the textual entity feature, representing $N$ aggregated entity-clause features from $N$ channels of $\mathbf{a}_q$.

For the reference image, we use a similar aggregation process, but with the global features of the reference image instead of the summary feature. Specifically, we also utilize learnable queries $\mathbf{b}_q = {b_1, ..., b_k}$ and use the local features $\textbf{E}^{l}_r$ and global features $\textbf{E}^{g}_r$ of the reference image as inputs to the transformer, adaptively aggregating corresponding feature channels for the same visual entity, formulated as follows,
\begin{equation}
        \hat{\mathbf{b}}_q = \operatorname{Transformer}\left(\left[\textbf{E}^{g}_r, \textbf{E}^{l}_r, \mathbf{b}_q\right]\right),
        \label{eq:bq}
\end{equation}
where $\hat{\mathbf{b}}_q\!\!\in\!\mathbb{R}^{N\times D}$ is the visual entity feature.

\noindent \textbf{Multimodal Query Composition.}
So far, we have obtained the textual and visual entity features. To improve the model's multi-granularity perception of multimodal queries, we concatenate these features with the global features of the reference image and the MMT, resulting in the final entity feature. For the MMT, the final entity feature is $\hat{\textbf{E}}_m = [\textbf{E}^{g}_m, \hat{\mathbf{a}}_q] \in \mathbb{R}^{(1+N) \times D}$. For the reference image, it is $\hat{\textbf{E}}_r = [\textbf{E}^{g}_r, \hat{\mathbf{b}}_q] \in \mathbb{R}^{(1+N) \times D}$, where $N$ is the number of channels in the learnable queries.

Then, following previous CIR methods~\cite{tgcir, blip4cir}, we use the same composition module for multimodal query features $\hat{\textbf{E}}_m$ and $\hat{\textbf{E}}_r$ to get composed feature $\textbf{E}_c$. Finally, we introduce the loss functions (including the summary-guided distillation strategy, orthogonal regularization, and batch-based classification loss) and train-inference phases of TEMA in \textbf{Appendix}~\ref{sup:train_strategy}. And we represented the algorithm of TEMA's processing flow in \textbf{Appendix}~\ref{sup:algorithm}.

%% file: 2-sec/4_exp.tex
\section{Experiments}
\label{exp}
In this section, we discuss the detailed experiments.

\begin{table*}[ht]
  \centering
  \caption{Performance comparison on M-FashionIQ and M-CIRR relative to R@K(\%). The overall best results are in bold, while best results over baselines are underlined. The Avg metric in M-CIRR denotes (R@$5$ + R$_{subset}$@$1$) / 2. }
  \vspace{-8pt}
        \resizebox{\linewidth}{!}{
        
    \begin{tabular}{l|c|cc|cc|cc|cc|ccc|cc|c}
        \Xhline{2pt}
    \hline
    \multicolumn{1}{c|}{\multirow{3}{*}{Method}}& \multirow{3}{*}{Text Encoder} & \multicolumn{8}{c|}{M-FashionIQ}                              & \multicolumn{6}{c}{M-CIRR} \\
\cline{3-16}      &        & \multicolumn{2}{c|}{Dresses} & \multicolumn{2}{c|}{Shirts} & \multicolumn{2}{c|}{Tops\&Tees} & \multicolumn{2}{c|}{Avg} & \multicolumn{3}{c|}{R@k} & \multicolumn{2}{c|}{R$_{subset}$@k} & \multirow{2}{*}{Avg} \\
\cline{3-15}       &      & R@10  & R@50  & R@10  & R@50  & R@10  & R@50  & R@10 & R@50  & k=1   & k=5   & k=10  & k=1   & k=2   &  \\
    \hline
    \hline
    Text-only & BLIP      & 24.72  & 47.94  & 26.88  & 49.67  & 27.25  & 50.22  & 26.28  & 49.28  & 23.50  & 50.10  & 67.30  & 49.28  & 67.70  & 49.69  \\
    Text+Image & BLIP      & 38.96  & 60.57  & 34.48  & 56.39  & 41.50  & 63.13  & 38.31  & 60.03  & 36.46  & 70.94  & 81.57  & 64.73  & 80.53  & 67.84  \\
        \rowcolor[rgb]{ .949,  .949,  .949} \multicolumn{16}{c}{\textit{Traditional Model-Based Methods}}\\
    TIRG& LSTM      & 7.88  & 14.35  & 9.66  & 18.30  & 10.07  & 21.51  & 9.20  & 18.05  & 9.68  & 30.73  & 47.81  & 14.92  & 36.31  & 22.83  \\
    CLVC-Net& LSTM      & 14.86  & 27.55  & 16.18  & 31.05  & 17.14  & 34.27  & 16.06  & 30.96  & 11.60  & 36.22  & 50.26  & 19.67  & 40.82  & 27.95  \\
    FashionViL& BERT  & 22.07      & 47.81      & 22.32      & 46.93      & 29.08      & 55.62      & 24.49  & 50.12  & -     & -      & -      & -      & -      & - \\
    MGUR&  RoBERTa     & 21.42  & 45.27  & 16.58  & 37.59  & 23.92  & 49.16  & 20.64  & 44.01  & -     & -     & -     & -     & -     & - \\
    \rowcolor[rgb]{ .949,  .949,  .949} \multicolumn{16}{c}{\textit{VLP Model-Based Methods}}\\
    FashionSAP& ALBEF  & 25.63  & 51.81  & 26.89  & 51.82  & 32.33  & 59.97  & 28.28  & 54.53  & -     & -     & -     & -     & -     & - \\
    BLIP4CIR& BLIP  & 40.97  & 63.28  & 37.81  & 59.10  & 44.19  & 64.94  & 40.99  & 62.44  & 39.92   & 74.04  & 84.07  & 67.79      & 82.21      & 70.92 \\
    BLIP4CIR+Bi& BLIP  & 41.51  & 63.23  & 37.51  & 57.92  & 43.32  & 65.00  & 40.78  & 62.05  & 41.24 & 75.61  & 84.21 & 69.46      & 82.89      & 72.54 \\
    Candidate& BLIP  & \underline{43.30}  & \underline{65.36}  & \underline{47.96}  & \underline{65.53}  & \underline{50.87}  & \underline{69.23}  & \underline{47.38}  & \underline{66.71}  & \underline{42.03}      & \underline{75.92} & \underline{84.61}   & \underline{69.58}      & \underline{83.84}   & \underline{72.75} \\
    \hline
    \hline
            \rowcolor[rgb]{ .851,  .851,  .851}
    \multicolumn{1}{l}{\textbf{TEMA~(Ours)}} &    BLIP   & \textbf{45.74} & \textbf{69.48} & \textbf{50.35} & \textbf{71.26} & \textbf{55.67} & \textbf{75.52} & \textbf{50.59} & \textbf{72.09} & \textbf{45.29} & \textbf{79.46} & \textbf{88.17} & \textbf{72.05} & \textbf{86.52} & \textbf{75.76} \\
    \hline
        \Xhline{2pt}
    \end{tabular}%
    }
  \label{tab:main}%
\end{table*}%

\subsection{Experimental Settings}
\textbf{Evaluation.} We use our proposed multi-modification datasets for training and evaluation, while choosing the recall at rank K (R@K) as the evaluation metric, quite similar to the previous CIR task~\cite{val}. The datasets include a fashion-domain dataset, M-FashionIQ, and an open-domain dataset, M-CIRR. For the evaluation of both, we use the validation splits. 
For M-FashionIQ, we employ R@$10$, R@$50$ and their category-wise averages. And M-CIRR assessment included R@$k$ ($k\!\!=\!\! 1, 5, 10$), R$_{subset}@k$ ($k\!\!=\!\! 1, 2$) and the average (R@$5$ + R$subset@1$) / $2$. In addition, we provide a detailed description of the above datasets (detailed in \textbf{Appendix}~\ref{sup:datasets}).

\noindent \textbf{Implementation Details.} We utilize BLIP~\cite{blip} as the backbone and freeze the image encoder. We train TEMA using the AdamW optimizer with an initial learning rate of $2e$-$5$. The batch size is set to $64$, and we maintain a dimension of $256$. The channel number $N$ of the learnable queries is set to $3$ for both M-FashionIQ and M-CIRR. Through a simple grid search, we set $\kappa$ to $0.6$ and $\mu$ to $0.2$ in Eqn~(\ref{eq:optimization}). 
All experiments are accomplished on a single NVIDIA A40 GPU with 48 GB memory.

\subsection{Method Comparison}
We conducted a comprehensive evaluation of the proposed TEMA model against several significant baselines using the two constructed datasets, \textit{i.e.}, M-FashionIQ and M-CIRR. We also provide results on traditional CIR benchmarks in \textbf{Appendix}~\ref{sup:traditional_cir}. Since the source codes for some methods were not accessible, we selected several open-source baselines (MGUR~\cite{mgur}, BLIP4CIR~\cite{blip4cir}, Candidate~\cite{candidate}, etc.). We retrained and tested them according to their original settings on two multi-modification datasets. It is important to note that, due to the token length limitations of the CLIP text encoder, we did not use it as a backbone. The results are presented in Table~\ref{tab:main}, leading to the following conclusions:
1) Our proposed TEMA achieves superior performance on both M-FashionIQ and M-CIRR, indicating its excellent generalization capabilities and robust comprehension of queries in both fashion and open-domain contexts.
2) TEMA demonstrates a significant performance advantage over the baselines, which also utilize BLIP as their backbone. This superiority may be attributed to the enhancements provided by our PA and EM modules, which improve the model's ability to grasp the nuances of MMT.
3) The performance of BLIP-based models markedly surpasses that of models employing traditional backbones, suggesting that, compared to conventional architectures (such as ResNet and LSTM), VLP-based models are more adept at understanding complex MMT.

\subsection{Ablation Study}
\label{sec:abla}
In this section, we introduce the ablation study of our proposed TEMA with different variants, as shown in Table~\ref{tab:abla}. The compared variants are as follows.
\textbf{$\bullet~$w/o PA.} We train TEMA without the PA module.
\textbf{$\bullet~$w/o CD.} We only ablate the Consistency Detector (CD) of PA, i.e., we don't check the summary generated by LLM.
\textbf{$\bullet~$w/o EM, w/o EM\_txt, and w/o EM\_img.} We first remove the entire EM, using the summary for composition instead, i.e., w/o EM. Further, we perform EM only for one modality, i.e., w/o EM\_txt and w/o EM\_img.
\textbf{$\bullet~$w/o Summ.} We don't use the Summary-guided Distillation strategy in this setup. Instead we only perform the other two losses.
\textbf{$\bullet~$w/o Ortho, w/o Ortho\_txt, and w/o Ortho\_img.} To investigate the role of Orthogonal Regularization for entity features, we removed it for both the textual entity feature and visual entity feature. Furthermore, we investigate Orthogonal Regularization by only performing for visual or textual entity features.

\begin{table}[t!]
  \centering
  \caption{Ablation study on M-FashionIQ and M-CIRR datasets. We compute Avg-R@10, R@50 for M-FashionIQ, and Avg (mean of R@$5$ and R$_{subset}$@$1$) for M-CIRR, respectively.}
\resizebox{\linewidth}{!}{
    \begin{tabular}{cc|cccccc}
    \Xhline{2pt}
    \multicolumn{2}{c|}{\multirow{2}{*}{Method}} & \multicolumn{4}{c|}{M-FashionIQ} & \multicolumn{2}{c}{M-CIRR} \\
\cline{3-8}    \multicolumn{2}{c|}{} & R@$10$ & \multicolumn{1}{c|}{$\Delta$} & R@$50$ & \multicolumn{1}{c|}{$\Delta$} & Avg   & $\Delta$ \\
    \hline
    \hline
            \rowcolor[rgb]{ .949,  .949,  .949} \multicolumn{8}{c}{\textit{MMT Parsing Assistant (PA)}}\\
    \multicolumn{2}{l|}{w/o PA} &  47.80     &\multicolumn{1}{c|}{\cellcolor[rgb]{ 0.573,  0.816,  0.314}-2.79}       &  69.83     &\multicolumn{1}{c|}{\cellcolor[rgb]{ 0.573,  0.816,  0.314}-2.26}       &71.59       &\cellcolor[rgb]{ 0.573,  0.816,  0.314}-4.17  \\
        \multicolumn{2}{l|}{w/o CD} &49.14       &\multicolumn{1}{c|}{\cellcolor[rgb]{ 0.82,  0.902,  0.710}-1.45}       &70.96       &\multicolumn{1}{c|}{\cellcolor[rgb]{ 0.82,  0.902,  0.710}-1.13}       &73.87       &\cellcolor[rgb]{ 0.82,  0.902,  0.710}-1.89  \\
                \rowcolor[rgb]{ .949,  .949,  .949} \multicolumn{8}{c}{\textit{MMT-oriented Entity Mapping (EM)}}\\
    \multicolumn{2}{l|}{w/o EM} & 45.41      &\multicolumn{1}{c|}{\cellcolor[rgb]{ 0.573,  0.816,  0.314}-5.18}       & 68.18      &\multicolumn{1}{c|}{\cellcolor[rgb]{ 0.573,  0.816,  0.314}-3.91}       & 70.99      &\cellcolor[rgb]{ 0.573,  0.816,  0.314}-4.77  \\
    \multicolumn{2}{l|}{w/o EM\_txt}  & 46.11      &\multicolumn{1}{c|}{\cellcolor[rgb]{ 0.82,  0.902,  0.710}-4.48}       & 68.25      &\multicolumn{1}{c|}{\cellcolor[rgb]{ 0.82,  0.902,  0.710}-3.84}       & 71.20      &\cellcolor[rgb]{ 0.82,  0.902,  0.710}-4.56  \\
    \multicolumn{2}{l|}{w/o EM\_img}&46.17       &\multicolumn{1}{c|}{\cellcolor[rgb]{ 0.910,  0.953,  0.855}-4.42}       &68.72       &\multicolumn{1}{c|}{\cellcolor[rgb]{ 0.910,  0.953,  0.855}-3.37}       & 71.64      &\cellcolor[rgb]{ 0.910,  0.953,  0.855}-4.12  \\
                \rowcolor[rgb]{ .949,  .949,  .949} \multicolumn{8}{c}{\textit{Loss Functions}}\\

    \multicolumn{2}{l|}{w/o Summ} & 49.40 & \multicolumn{1}{c|}{\cellcolor[rgb]{ 0.910,  0.953,  0.855}-1.19} & 71.14 & \multicolumn{1}{c|}{\cellcolor[rgb]{ 0.82,  0.902,  0.710}-0.95} & 74.16
    &\cellcolor[rgb]{ 0.82,  0.902,  0.710}-1.60 \\
    \multicolumn{2}{l|}{w/o Ortho} & 49.38 & \multicolumn{1}{c|}{\cellcolor[rgb]{ 0.82,  0.902,  0.710}-1.21} & 71.58 & \multicolumn{1}{c|}{\cellcolor[rgb]{ 0.910,  0.953,  0.855}-0.51} & 75.02    & \cellcolor[rgb]{ 0.910,  0.953,  0.855}-0.74 \\
    \multicolumn{2}{l|}{w/o Ortho\_txt} & 48.14      &\multicolumn{1}{c|}{\cellcolor[rgb]{ 0.725,  0.855,  0.561}-2.45}       & 69.96     &\multicolumn{1}{c|}{\cellcolor[rgb]{ 0.725,  0.855,  0.561}-2.13}       & 73.39      &\cellcolor[rgb]{ 0.725,  0.855,  0.561}-2.37  \\
    \multicolumn{2}{l|}{w/o Ortho\_img} & 48.93      &\multicolumn{1}{c|}{\cellcolor[rgb]{ 0.713,  0.903,  0.648}-1.66}       & 70.44      &\multicolumn{1}{c|}{\cellcolor[rgb]{ 0.713,  0.903,  0.648}-1.65}       & 73.61      &\cellcolor[rgb]{ 0.713,  0.903,  0.648}-2.15  \\
    \hline
    \hline
    \multicolumn{2}{l|}{\textbf{TEMA}} 
    &  \textbf{50.59}
    & \multicolumn{1}{c|}{\cellcolor[rgb]{ 1,  0.983,  0.717}   -0.00} 
    &  \textbf{72.09}     
    & \multicolumn{1}{c|}{\cellcolor[rgb]{ 1,  0.983,  0.717}   -0.00}          
    &  \textbf{75.76}      
    & \cellcolor[rgb]{ 1,  0.983,  0.717}   -0.00         \\
    \Xhline{2pt}
    \end{tabular}%
    }
      \vspace{-10pt}
  \label{tab:abla}%
\end{table}%

From the ablation results of TEMA in Table~\ref{tab:abla}, we have following four observations. 1) Both w/o PA and w/o CD are inferior to TEMA. 
In particular, removing PA causes very substantial performance degradation. This is reasonable that the PA module is a powerful training aid, and the summary generated by the PA module serves as a guide for the textual entity feature and facilitates the training of the EM module to aggregate the visual entities and multiple clauses within MMT, respectively, demonstrating the importance of using the MMT Parsing Assistant. 
2) w/o EM, w/o EM\_txt, and w/o EM\_img all perform worse than TEMA, and removing any of the components of EM resulted in a more substantial drop than the other module ablations, indicating that the entity mapping process indeed improved the MMT comprehension for TEMA, by aggregating complex modification clauses to several to-be-modified entities.
3) Both w/o Summ and w/o Ortho are inferior to TEMA, showing their necessity in TEMA's optimization. 
4) In w/o Ortho\_txt and w/o Ortho\_img, the orthogonal regularization is performed on one modality but causes more drastic performance degradation than w/o Ortho. This may be because such an asymmetric process destroys the alignment of the features of two modalities. We also provide more detailed ablation results in \textbf{Appendix}~\ref{sup:ablation_study}.

\subsection{Performance on Traditional CIR}
To verify the performance of TEMA on traditional CIR, we conducted additional experiments, training and testing TEMA in the settings of original FashionIQ and CIRR datasets. The performance comparison is shown in Table~\ref{tab:traditional_cir}.
We can observe that TEMA's performance is better than the previous baselines, which proves the powerful generalization ability of TEMA. It not only performs well on the multi-modification CIR benchmark, but also maintains the performance on CIR. Additionally, we represented the full results in \textbf{Appendix}~\ref{sup:traditional_cir}.
\begin{table}[h]
  \centering
  \caption{Performance on traditional CIR benchmarks, including FashionIQ and CIRR.} 
  \resizebox{0.8\linewidth}{!}{
    \begin{tabular}{cc|cc|cccccc}
    \Xhline{2pt}
    \multicolumn{2}{c|}{\multirow{2}{*}{Methods}}&\multicolumn{2}{c|}{\multirow{2}{*}{Backbone}} & \multicolumn{4}{c|}{FashionIQ} & \multicolumn{2}{c}{CIRR} \\
\cline{5-10}    \multicolumn{2}{c|}{}& \multicolumn{2}{c|}{} & \multicolumn{2}{c}{R@$10$} & \multicolumn{2}{c|}{R@$50$} & \multicolumn{2}{c}{Avg}  \\
    \hline
    \hline

    \multicolumn{2}{l|}{CASE}
    &  \multicolumn{2}{c|}{BLIP}
    &  \multicolumn{2}{c}{48.79}
    &  \multicolumn{2}{c|}{70.68}     
    &  \multicolumn{2}{c}{77.50}      
         \\ 
    \multicolumn{2}{l|}{Candidate}
    &  \multicolumn{2}{c|}{BLIP}
    &  \multicolumn{2}{c}{51.17}
    &  \multicolumn{2}{c|}{73.13}     
    &  \multicolumn{2}{c}{\textbf{80.90}}       
         \\

    \multicolumn{2}{l|}{CoVR-BLIP}
    &  \multicolumn{2}{c|}{BLIP}
    &  \multicolumn{2}{c}{48.53}
    &  \multicolumn{2}{c|}{70.25}     
    &  \multicolumn{2}{c}{76.81}     
         \\
         \rowcolor[rgb]{ .851,  .851,  .851}
    \multicolumn{2}{l|}{\textbf{TEMA}}
    &  \multicolumn{2}{c|}{BLIP}
    &  \multicolumn{2}{c}{\textbf{53.02}}
    &  \multicolumn{2}{c|}{\textbf{74.20}}     
    &  \multicolumn{2}{c}{{80.18}} 
    \\
         
    \Xhline{2pt}
    \end{tabular}%
    }
\label{tab:traditional_cir}
    \end{table}
\subsection{Sensitivity Analysis}
\label{sec:sensi}
\begin{figure}[t]
  \centering
  \includegraphics[width=\linewidth]{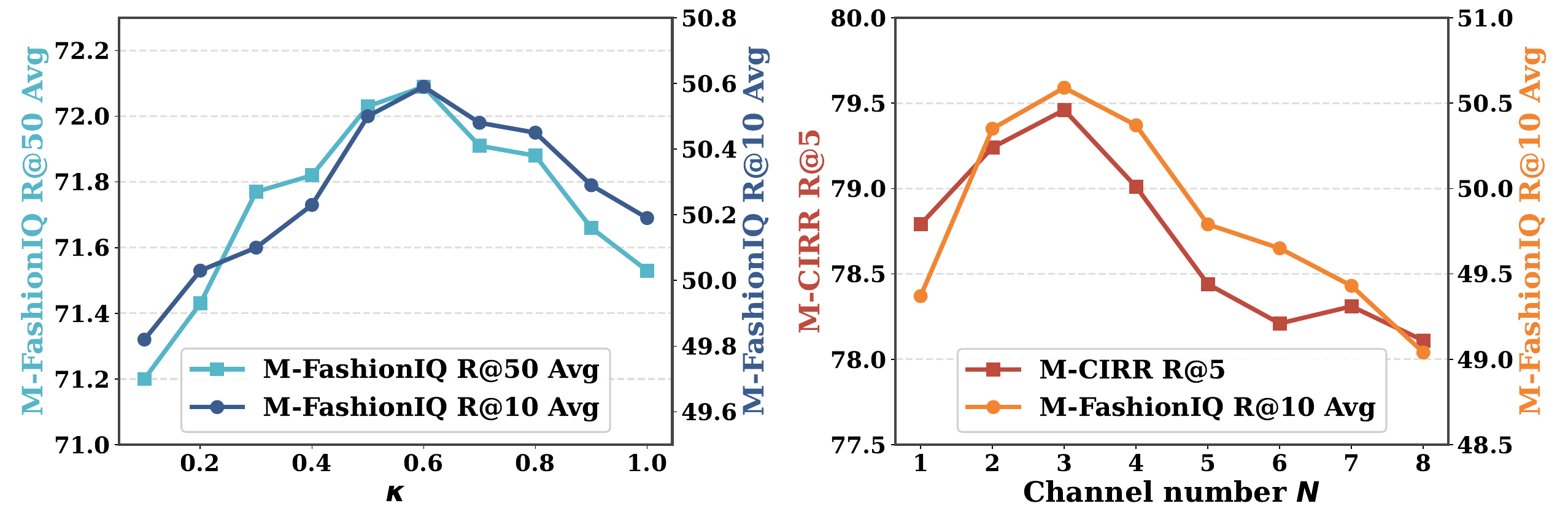}
\vspace{-18pt}
  \caption{Sensitivity analysis on the hyper-parameter $\kappa$ and the channel number $N$ of learnable queries.}
\vspace{-16pt}
  \label{fig:sensi}
\end{figure}

As shown in Figure~\ref{fig:sensi}, we evaluate the sensitivity of our proposed TEMA regarding the hyper-parameter $\kappa$ in Eqn~(\ref{eq:optimization}) on M-FashionIQ, and the channel number $N$ of learnable queries on both M-FashionIQ and M-CIRR.
As the results show, the performance of TEMA initially improves with the increase of $\kappa$, reaching an optimal level, after which it gradually declines as $\kappa$ continues to rise. This is reasonable, as the summary-guided distillation loss $\mathcal{L}_{summ}$ requires a certain weight to enhance the optimization effect. When the value is too high, it may lead to an imbalance among different loss functions.
For the channel number of learnable queries, denoted by $N$, the performance of TEMA first shows an upward trend on both M-FashionIQ and M-CIRR, reaching the optimal. This is because a certain number of channels are needed to correspond to different to-be-modified entities. However, when the value of 
$N$ becomes too large, performance begins to fluctuate and decline, as an excess of channels may lead to confusion among the to-be-modified entities, thereby reducing retrieval performance.

\subsection{Qualitative results for PA module}
\label{sec:pa_qua}
\begin{figure}[h]
\includegraphics[width=\linewidth]{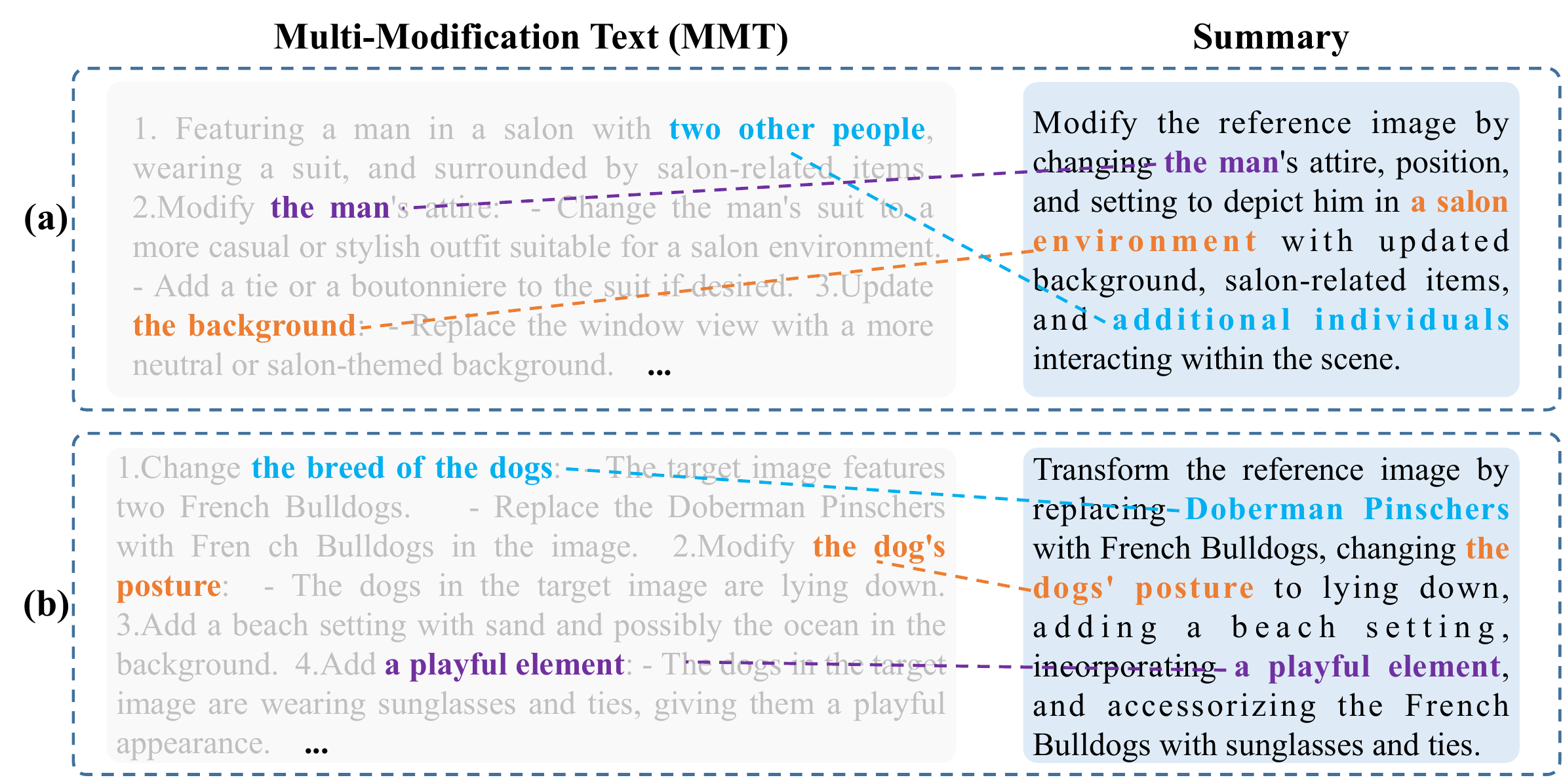}
\vspace{-15pt}
  \caption{The qualitative results for the PA module, which shows the MMT and corresponding summary. The to-be-modified entities are colored.}
\vspace{-18pt}
  \label{fig:pa_qua}
\end{figure}

To investigate the validity of the PA-generated summaries, we present qualitative results from the MMT Parsing Assistant (PA) to verify whether the summary includes all the to-be-modified entities. 
As shown in Figure~\ref{fig:pa_qua}, we highlight these entities in the MMT, where they appear as subjects in clauses.
The summary generated by PA captures all to-be-modified entities in the MMT. For example, in Figure~\ref{fig:pa_qua}(b), the summary accurately identifies entities such as ``the breed of the dogs'' and ``the dog's posture''.
The key difference from the MMT is the omission of certain detailed descriptions, which condenses the focus on the to-be-modified entities.
This approach effectively guides the model, helping it identify the entities while minimizing distractions from lengthy descriptions, conjunctions, and prepositions in the MMT. Consequently, this enhances the model and facilitates the subsequent EM module in aggregating the to-be-modified entities.
Additionally, we provide attention visualization results on the summaries in \textbf{Appendix}~\ref{sup:summary_vis} and more qualitative results in \textbf{Appendix}~\ref{sup:quali}.

%% file: 2-sec/5_conclusion.tex
\section{Conclusion}
In this work, we addressed two limitations that were highly relevant to CIR's practical applications, namely Insufficient Entity Coverage and Clause-Entity Misalignment, thereby advancing CIR toward real-world use. We constructed two multi-modification datasets, M-FashionIQ and M-CIRR. In addition, we proposed TEMA, which was the first CIR framework designed for multi-modification while also accommodating simple modifications. TEMA outperformed previous methods in both original and multi-modification scenarios, showcasing its superiority.

%% file: 2-sec/6_limitations.tex
\section{Limitations}
Although this study constructs the M-FashionIQ and M-CIRR datasets and proposes the TEMA framework, which, through multi-modification text (MMT) parsing and entity mapping, achieves substantial progress in composed image retrieval (CIR) across original and multi-modification scenarios, several limitations remain.
First, unlike other data used for pretraining, our CIR datasets for multi-modification scenarios are designed to provide a training and evaluation environment that is closer to real applications, rather than to solely increase model performance. Because the annotations in the constructed datasets are longer, they increase the difficulty for models to understand modification intentions and therefore do not necessarily lead to higher retrieval metrics.
Second, the PA module incorporates large language models during training. Although this module is disabled during testing, it still introduces minor computational overhead in training.
Finally, consistent with most current CIR studies, the proposed TEMA currently supports only single turn retrieval, and its effectiveness in multi turn interactive CIR scenarios remains to be explored.
Future research should address these limitations to enhance the practical utility of our proposed datasets and TEMA model.

%% file: 2-sec/7_Ethical.tex
\section{Ethical Considerations}
First, for the constructed datasets, our public release will remove personally identifiable information and prohibits retrieval based on identifiable faces; the license explicitly disallows surveillance uses. We encourage deployments to incorporate abuse detection, rate limiting, keyword and category block lists, and context aware access control policies, and to state permitted uses and prohibitions at the time of releasing data and models.
Second, composed image retrieval (CIR) could be repurposed for sensitive settings. We will decline data requests that target such settings and will provide an email address to process takedown requests for problematic content.
Finally, we plan to release TEMA's code, prompts, and datasets under a research only license in order to minimize misuse.
Overall, we will continue to validate and improve this work through careful safety and compliance practices across broader populations and scenarios, thereby ensuring a responsible contribution to the CIR community.

%% file: 2-sec/supplementary.tex
\clearpage
\appendix
\setcounter{page}{1}
\noindent This is the appendix of ``TEMA: Anchor the Image, Follow the Text for Multi-Modification
Composed Image Retrieval''. 
\begin{itemize}
    \item \textbf{Appendix~\ref{sup:datasets}}: Multi-Modification Datasets
    \begin{itemize}
        \item \textbf{Appendix~\ref{sup:data_stat}}: Dataset Statistics
        \item \textbf{Appendix~\ref{sup:metrics}}: Metrics
    \end{itemize}
    \item 
    \textbf{Appendix~\ref{sup:data_cons}}: 
    More Details for Dataset Construction
    \begin{itemize}
        \item \textbf{Appendix~\ref{sup:quality_check}}: Quality Check
        \item 
        \textbf{Appendix~\ref{sup:content_filter}}: Content Filter
    \end{itemize}
    \item \textbf{Appendix~\ref{sup:train_strategy}}: Training Strategy
    \begin{itemize}
        \item \textbf{Appendix~\ref{sup:loss_functions}}: Loss Functions
        \item 
        \textbf{Appendix~\ref{sup:train_inference}}: Train and Inference
    \end{itemize}
    \item \textbf{Appendix~\ref{sup:more_quantitative}}: More Quantitative Results
    \begin{itemize}
        \item \textbf{Appendix~\ref{sup:ablation_study}}: Ablation Study
        \item \textbf{Appendix~\ref{sup:computation_cost}}: Computation Cost Analysis
        \item \textbf{Appendix~\ref{sup:traditional_cir}}: Evaluation on Traditional CIR Benchmarks
    \end{itemize}
    \item \textbf{Appendix~\ref{sup:algorithm}}: Algorithm of TEMA's Training and Inference Process
    \item \textbf{Appendix~\ref{sup:prompt}}: More Analysis on Prompts
    \begin{itemize}
        \item \textbf{Appendix~\ref{sup:detailed_prompts}}: Detailed Prompts for MMT Generation
        \item \textbf{Appendix~\ref{sup:analysis_different_prompts}}: Analysis of Different Prompts
    \end{itemize}
    \item \textbf{Appendix~\ref{sup:more_qualitive_results}}: More Qualitative Results
    \begin{itemize}
        \item \textbf{Appendix~\ref{sup:false_negative}}: Mitigation on False-negative Samples
    \item \textbf{Appendix~\ref{sup:summary_vis}}: Attention Visualization for Summaries
    \item \textbf{Appendix~\ref{sup:quali}}: Case Study
    \end{itemize}

\end{itemize}

\section{Multi-Modification Datasets}
\label{sup:datasets}
To evaluate the validity of models in multi-modification scenarios, we constructed two datasets. We now describe each dataset in detail as follows,
\begin{itemize}
    \item \textbf{M-FashionIQ} is based on the classic CIR dataset, the \textbf{FashionIQ}~\cite{FashionIQ}, whose content belongs entirely to the fashion domain. It consists of $77,684$ images which are divided into three categories: \textit{Dresses}, \textit{Shirts}, and \textit{Tops\&Tees}. Following the FashionIQ, we treat it as three independent datasets. Following previous CIR methods, we utilize \mbox{$\sim$\hspace{0em}$46$K} and \mbox{$\sim$\hspace{0em}$15$K} images for training and testing, respectively. Finally, there are $18$K triplets for training and \mbox{$\sim$\hspace{0em}$6$K} triplets for testing.
    \item \textbf{M-CIRR} is based on the classic open-domain CIR dataset, the \textbf{CIRR}~\cite{cirr}. It contains $21,552$ real images taken from the renowned language reasoning dataset ${\operatorname{NLVR}}^2$, which is well-known for its natural language reasoning applications. Specifically, we utilize \mbox{\hspace{0em}$28,225$} and \mbox{\hspace{0em}$4,181$} triplets for training and testing, respectively. In addition, following CIRR, M-CIRR includes a specialized subset designed for fine discrimination. This subset focuses on negative images that exhibit a high degree of visual similarity and is utilized to assess the model's performance in distinguishing false-negative images.
\end{itemize}

\begin{table*}[t!]
  \centering
  \caption{Comparison of modification text in lengths and attributes between the original CIR datasets and the expanded multi-modification datasets. The length is counted by tokens. In the attributes, \textbf{CR} represents Complex Relations (CR), \textbf{ME} denotes Multiple Entities (ME), and \textbf{FG} means Fine-Grained (FG).}
          \resizebox{0.6\linewidth}{!}{
    \begin{tabular}{l|cccccc}
    \hline
    \hline
    \multicolumn{1}{c|}{\multirow{2}{*}{Method}} & \multicolumn{3}{c|}{Length of Modification Text} & \multicolumn{3}{c}{Attribute}\\
\cline{2-7}          & \#Minimal & \#Maximal & \multicolumn{1}{c|}{\#Average} & \multicolumn{1}{c}{CR} & \multicolumn{1}{c}{ME} & \multicolumn{1}{c}{FG}\\
    \hline
    FashionIQ & 3.0     & 37.0    & \multicolumn{1}{c|}{24.7} & \multicolumn{1}{c}{\textcolor{red}{\ding{55}}}& \multicolumn{1}{c}{\textcolor{red}{\ding{55}}}& \multicolumn{1}{c}{\textcolor{red}{\ding{55}}}\\
            \rowcolor[rgb]{ .949,  .949,  .949}
    \textbf{M-FashionIQ} & 25.0    & 327.0   & \multicolumn{1}{c|}{152.7} &\multicolumn{1}{c}{\textcolor{green}{\ding{51}}}&\multicolumn{1}{c}{\textcolor{green}{\ding{51}}}&\multicolumn{1}{c}{\textcolor{green}{\ding{51}}}\\
    \hline
    CIRR  & 2.0     & 50.0    & \multicolumn{1}{c|}{12.8} & \multicolumn{1}{c}{\textcolor{red}{\ding{55}}}& \multicolumn{1}{c}{\textcolor{red}{\ding{55}}}& \multicolumn{1}{c}{\textcolor{green}{\ding{51}}}\\
            \rowcolor[rgb]{ .949,  .949,  .949}
    \textbf{M-CIRR} & 35.0    & 468.0   & \multicolumn{1}{c|}{319.4}& \multicolumn{1}{c}{\textcolor{green}{\ding{51}}}& \multicolumn{1}{c}{\textcolor{green}{\ding{51}}}& \multicolumn{1}{c}{\textcolor{green}{\ding{51}}}\\
    \hline
    \hline
    \end{tabular}%
    }
    \label{tab:token}
 \end{table*}%

\subsection{Dataset Statistics}
\label{sup:data_stat}
For evaluation consistency, we note that existing CIR works report validation set results for FashionIQ, and CIRR's test set ground truth is not publicly available. Therefore, we expand modification texts to MMT in both training and validation sets of FashionIQ and CIRR, combining them with the original reference and target images to create new triplet collections, which serve as the training and test sets for M-FashionIQ and M-CIRR, respectively.
Specifically, we process a total of $24,016$ queries in the FashionIQ dataset and $32,406$ queries in the CIRR dataset. As illustrated in Table~\ref{tab:token}, the minimum, maximum, and average lengths of the modification texts in our constructed M-FashionIQ and M-CIRR datasets significantly increase compared to the original datasets. 
This provides more room for modification texts. For example, M-FashionIQ and M-CIRR contain all three attributes of CR, ME, and FG, while the original FashionIQ does not have any of them. CIRR does not have CR and ME, and its dense labeling with FG is not widely used.
These more comprehensive and detailed descriptions better capture users' nuanced composed retrieval needs in real-world scenarios.

\subsection{Metrics}
\label{sup:metrics}
In terms of model evaluation, we train the model via the training sets of M-FashionIQ and M-CIRR, and then evaluate the model on the validation sets. Finally, we utilize the same evaluation metric, \textit{i.e.}, recall at rank $k$ (R@$k$), which is conventionally used in CIR tasks~\cite{FashionIQ, cirr, tgcir, clip4cir-v2, clip4cir-v3}.

\begin{figure}[t!]
  \centering
  \includegraphics[width=1.0\linewidth]{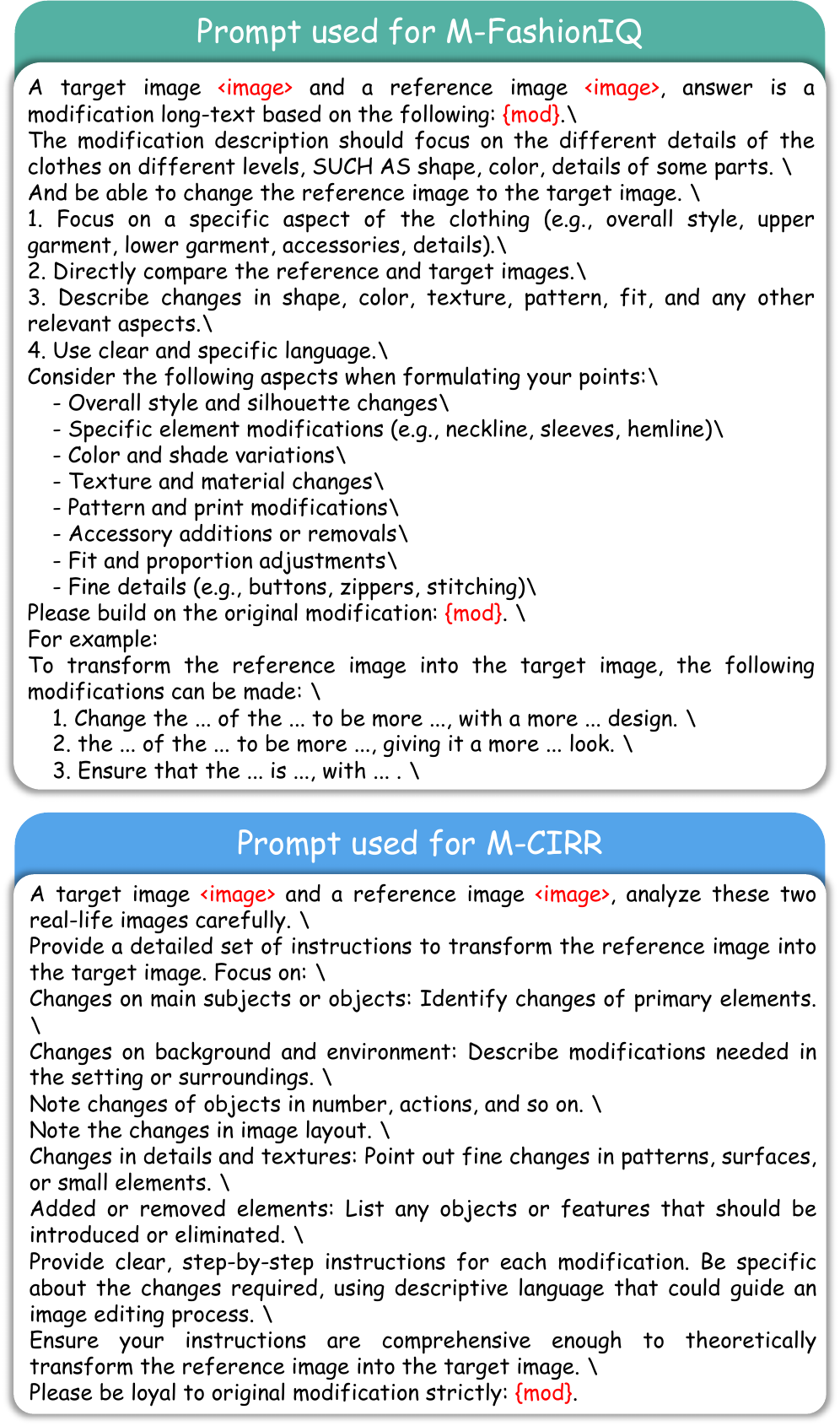}
  \caption{Prompts used in the process of MMT generation, for both M-FashionIQ and M-CIRR datasets.}

  \label{fig:sup_prompt}
\end{figure}

\section{More Details for Dataset Construction}
\label{sup:data_cons}
We supplemented more details about the dataset construction procedure, including Quality Chcek and Content Filter in Sec~\ref{sec:data_cons}.
\subsection{Quality Check}
\label{sup:quality_check}
During the Quality Check process in Sec~\ref{sec:data_cons}, we examine and revise the texts from four perspectives: 
\textbf{1) Consistency.} 
The modification text should describe plausible changes to objects or attributes within a given visual scene. Content that refers to irrelevant aspects, such as low-level image parameters (e.g., exposure, white balance), is considered inconsistent and should be removed. Annotators are responsible for ensuring semantic consistency within each text.
\textbf{2) Accuracy.}
The modification should accurately reflect the intended change without introducing speculative or hallucinated content. Annotators are required to verify that the described objects, attributes, or actions are plausible and grounded in real-world context, avoiding exaggeration or factual errors.
\textbf{3) Diversity.} 
To enhance the expressiveness and coverage of the dataset and better accommodate diverse user intents, the modification texts should exhibit linguistic and conceptual diversity. Annotators are encouraged to avoid repetitive sentence structures and instead adopt varied phrasings and perspectives to enrich the overall corpus.
\textbf{4) Quality.} 
As the initial MMTs are generated by an MLLM, they may suffer from issues such as unnatural phrasing or incoherent logic. Annotators are expected to refine and polish the texts where necessary, while preserving the original intent. A high-quality MMT should be fluent, precise, and reliable for both training and evaluation purposes.

\subsection{Content Filter}
\label{sup:content_filter}
Intuitively, even a well-formed modification text would be inappropriate if it fails to reflect a change relative to the reference image, e.g., describing attributes of the target image in isolation. Such cases deviate from the core principle of multi-modification annotations, where the modification should be conditioned on the reference image.
To address this issue, we introduce a \textit{Content Filter} stage following the manual refinement. Specifically, we feed the human-corrected MMTs along with the target image into a Large Language Model (LLM) to detect statements that directly describe the target image content without referencing the reference image. 
These cases indicate a breakdown in the referential grounding and render the reference image ineffective under the multi-modification formulation. We remove such statements from the MMTs to ensure that both the reference image and the MMT collaboratively contribute to the retrieval query.

\section{Training Strategy}
\label{sup:train_strategy}
\subsection{Loss Functions}
\label{sup:loss_functions}
\textbf{Summary-guided Distillation.}
Furthermore, to ensure that the text tokens generated by the EM module contain all the information about the to-be-modified entities, we employ a summary-guided distillation strategy that aligns the EM module's output with entities parsed by PA.
Specifically, for the textual entity feature $\hat{\mathbf{a}}_q$ in Eqn~(\ref{eq:aq}), we employ a simple cosine loss to close the semantics between the summary feature $\textbf{E}_s$ in Eqn~(\ref{eq:summ}) and it, formulated as follows,
\begin{equation}
    \mathcal{L}_{summ} = \operatorname{1} - \operatorname{cos}\left( \textbf{E}_s, \bar{\mathbf{a}}_q\right),
    \label{eq:cos}
\end{equation}
where $\bar{\mathbf{a}}_q$ indicates the average-pooled $\hat{\mathbf{a}}_q$.

\noindent \textbf{Orthogonal Regularization.} 
Considering if the $N$ channels of the entity feature accurately represent the aggregation of different to-be-modified entities, they should be semantically independent and orthogonal.
Inspired by TG-CIR~\cite{tgcir}, we design an Orthogonal Regularization to minimize the potential semantic overlap between channels, ensuring semantic independence.
\begin{equation}
    \mathcal{L}_{ortho} =\left\| {\hat{\mathbf{a}}_q}^{\top}\hat{\mathbf{a}}_q - \mathbf{I} \right\|_{F}^{2} + 
    \left\| {\hat{\mathbf{b}}_q}^{\top}\hat{\mathbf{b}}_q - \mathbf{I} \right\|_{F}^{2}, 
    \label{eq:ortho}
\end{equation}
where $\mathbf{I} \in \mathbb{R}^{N\times N}$ and $\left\|\cdot\right\|_F^2$ is the Frobenius norm of matrix. 

\noindent \textbf{Batch-based Classification Loss.} We apply the universal batch-based classification loss~\cite{val}, which serves as a variant of cross-entropy, to align $\textbf{E}_c$ with the target image feature $\textbf{E}_t$, formulated as follows,
\begin{equation}
\mathcal{L}_{bbc}\!\!=\!\! \frac{1}{B} \sum_{i=1}^{B} -\log \left\{ \frac{\exp \left\{ \operatorname{s} \left( \bar{\textbf{E}}_{ci} , \bar{\textbf{E}}_{ti} \right)  / \tau\right\}}{ \sum_{j=1}^{B} \exp \left\{ \operatorname{s} \left( \bar{\textbf{E}}_{ci}, \bar{\textbf{E}}_{tj} \right) / \tau \right\}  } \right\},
\label{bbc}
\end{equation}
where as $\bar{\textbf{E}}_{ci}, \bar{\textbf{E}}_{ti}$ indicate the average pooled $\textbf{E}_c, \textbf{E}_t$ of the $i$-th triplet, respectively. $\tau$ is the temperature coefficient. $B$ is the batch size.

\subsection{Train and Inference}
\label{sup:train_inference}
During training, we employ both the MMT Parsing Assistant (PA) and the MMT-oriented Entity Mapping (EM), and the final optimization function is formulated as,
\begin{equation}
    \mathbf{\Theta^{*}}=
    \underset{\mathbf{\Theta}}{\arg \min } \left( {\mathcal{L}}_{bbc} +\kappa {\mathcal{L}}_{summ}+ \mu {\mathcal{L}}_{ortho} \right),
    \label{eq:optimization}
    \end{equation}
where $\mathbf{\Theta^{*}}$ is the to-be-optimized parameter for TEMA and $\kappa, \mu$ are the trade-off hyper-parameters. 

During the inference phase, the MMT Parsing Assistant (PA) module is forbidden, while the MMT-oriented Entity Mapping (EM) module has learned how to understand the MMT from the PA module during training.
We fully compared the efficiency of our proposed TEMA with the SOTA of the CIR task.

\begin{table*}[ht]
  \centering
  \caption{Full validation results on traditional CIR benchmarks, including FashionIQ and CIRR.}
 \resizebox{\linewidth}{!}{
        
    \begin{tabular}{l|c|c|cc|cc|cc|cc|ccc|cc|c}
        \Xhline{2pt}
    \hline
    \multicolumn{1}{c|}{\multirow{3}{*}{Method}}& \multirow{3}{*}{Year} & \multirow{3}{*}{Text Encoder} & \multicolumn{8}{c|}{FashionIQ}                              & \multicolumn{6}{c}{CIRR} \\
\cline{4-17}      &        &       & \multicolumn{2}{c|}{Dresses} & \multicolumn{2}{c|}{Shirts} & \multicolumn{2}{c|}{Tops\&Tees} & \multicolumn{2}{c|}{Avg} & \multicolumn{3}{c|}{R@k} & \multicolumn{2}{c|}{R$_{subset}$@k} & \multirow{2}{*}{Avg} \\
\cline{4-16}       &      &       & R@10  & R@50  & R@10  & R@50  & R@10  & R@50  & R@10 & R@50  & k=1   & k=5   & k=10  & k=1   & k=2   &  \\
    \hline
    \hline
    \multicolumn{1}{l|}{CASE~\cite{case}} & 2024 &    BLIP   & 47.44 & 69.36 & 48.48 & 70.23 & 50.18 & 72.24 & 48.79 & 70.68 & 48.00 & 79.11 & 87.25 & 75.88 & 90.58 & 77.50 \\
    \multicolumn{1}{l|}{Candidate~\cite{candidate}} & 2024 &    BLIP   & 48.14 & 71.34 & 50.15 & 71.25 & 55.23 & 76.80 & 51.17 & 73.13 & \textbf{50.55} & 81.75 & \textbf{89.78} & \textbf{80.04} & \textbf{91.90} & \textbf{80.90} \\  
    \multicolumn{1}{l|}{CoVR-BLIP~\cite{covr}} & 2024 &    BLIP   & 44.55 & 69.03 & 48.43 & 67.42 & 52.60 & 74.31 & 48.53 & 70.25 & 49.69 & 78.60 & 86.77 & 75.01 & 88.12 & 76.81 \\  
    \rowcolor[rgb]{ .851,  .851,  .851}\multicolumn{1}{l}{\textbf{TEMA~(Ours)}} &  &    BLIP   & \textbf{49.66} & \textbf{71.98} & \textbf{52.90} & \textbf{73.55} & \textbf{56.49} & \textbf{77.07} & \textbf{53.02} & \textbf{74.20} & 49.15 & \textbf{82.18} & 88.81 & 78.17 & 90.32 & {80.18} \\
        \Xhline{2pt}
    \end{tabular}%
    }
\label{tab:full_traditional_cir}%
\end{table*}%

\section{More Quantitative Results}
\label{sup:more_quantitative}
\subsection{Ablation Study}
\label{sup:ablation_study}
To evaluate TEMA's generalization capability and the performance of widely accessible LLMs in multi-modification scenarios, we conducted additional ablation studies on the LLMs integrated into TEMA. As summarized in Table~\ref{tab:abla_llms}, we replaced the MLLM with other easily obtainable models, such as Qwen, LLaMA, and others, to generate MMT summaries. The results show that switching between different LLMs has minimal impact on TEMA's overall performance. 

This finding highlights TEMA's strong adaptability to various LLMs and demonstrates that it does not rely solely on proprietary models. Notably, TEMA maintains excellent training outcomes even when leveraging open-source models, such as LLaMA 3 or Qwen2-VL series, to replicate PA summarization and consistency checks. This reinforces the practicality and flexibility of TEMA in utilizing non-proprietary, openly available LLMs.
\begin{table}[t!]
  \centering
    \caption{Ablation study on different LLMs in TEMA.}
\resizebox{\linewidth}{!}{
    \begin{tabular}{cc|cccccc}
    \Xhline{2pt}
    \multicolumn{2}{c|}{\multirow{2}{*}{LLMs}} & \multicolumn{4}{c|}{M-FashionIQ} & \multicolumn{2}{c}{M-CIRR} \\
\cline{3-8}    \multicolumn{2}{c|}{} & \multicolumn{2}{c|}{R@$10$}  & \multicolumn{2}{c|}{R@$50$}  & \multicolumn{2}{c|}{Avg} \\
    \hline
    \hline
    \multicolumn{2}{l|}{gpt-4o-mini} 
    &  \multicolumn{2}{c|}{49.67}
    &  \multicolumn{2}{c|}{70.93} 
    &  \multicolumn{2}{c}{73.68}
         \\
    \multicolumn{2}{l|}{Qwen2-VL} 
    &  \multicolumn{2}{c|}{\textbf{51.12}}
    &  \multicolumn{2}{c|}{72.59}   
    &  \multicolumn{2}{c}{\textbf{75.83}}   
         \\
    \multicolumn{2}{l|}{Llama-2} 
    &  \multicolumn{2}{c|}{48.33}
    &  \multicolumn{2}{c|}{70.24}   
    &  \multicolumn{2}{c}{72.73}  
         \\
    \multicolumn{2}{l|}{Llama-3} 
    &  \multicolumn{2}{c|}{50.57}
    &  \multicolumn{2}{c|}{72.63}
    &  \multicolumn{2}{c}{75.31}
         \\

    \multicolumn{2}{l|}{Claude3.5-sonnet} 
    &  \multicolumn{2}{c|}{50.98}
    &  \multicolumn{2}{c|}{\textbf{73.02}}   
    &  \multicolumn{2}{c}{75.59}  
         \\        
    \Xhline{2pt}
    \end{tabular}%
    }
  \label{tab:abla_llms}%
  \vspace{-10pt}
\end{table}%

\subsection{Computation Cost Analysis}
\label{sup:computation_cost}
We determine the computational costs of our proposed TEMA compared to the sub-optimal model Candidate~\cite{candidate}. Specifically, we choose FLOPs, train time, test time, and GPU memory, as shown in Table~\ref{tab:cost}. All experiments were performed on a single NVIDIA A40 GPU and the batch size is set to $64$. 
The FLOPs represent the number of floating-point operations. The train time describes the time it takes for the model to optimize to the optimum, while the test time is the time it takes for the inference on one sample. It is worth noting that our proposed TEMA model is superior on all indicators, demonstrating its overall effectiveness.
\begin{table}[ht]
  \centering

  \caption{Comparison of TEMA and the sub-optimal model Candidate on computation cost. The better results are in bold.}
\resizebox{\linewidth}{!}{
    \begin{tabular}{cc|cc|cc|cc|cc|cc}
    \Xhline{2pt}
    \multicolumn{2}{c|}{Method} & \multicolumn{2}{c|}{Backbone} & \multicolumn{2}{c|}{FLOPs} & \multicolumn{2}{c|}{Train} & \multicolumn{2}{c|}{Test} & \multicolumn{2}{c}{Memory} \\
    \hline
    \multicolumn{2}{l|}{Candidate} & \multicolumn{2}{c|}{BLIP} & \multicolumn{2}{c|}{5.79G}  & \multicolumn{2}{c|}{16h} & \multicolumn{2}{c|}{16.7ms/sample} & \multicolumn{2}{c}{47.3G} \\
                \rowcolor[rgb]{ .851,  .851,  .851}
    \multicolumn{2}{c|}{\textbf{TEMA(Ours)}} & \multicolumn{2}{c|}{BLIP} & \multicolumn{2}{c|}{\textbf{3.68G}}  & \multicolumn{2}{c|}{\textbf{2.83h}} &  \multicolumn{2}{c|}{\textbf{7.9ms/sample}} & \multicolumn{2}{c}{\textbf{43.9G}} \\
    \Xhline{2pt}
    \end{tabular}%
    }

  \label{tab:cost}%
\end{table}%

\subsection{Evaluation on Traditional CIR Benchmarks}
\label{sup:traditional_cir}
In Table~\ref{tab:full_traditional_cir}, we supplemented TEMA's performance on the original CIR datasets (FashionIQ and CIRR). The results show that TEMA is superior to existing SOTA on most metrics, sufficiently demonstrating TEMA's extensibility.

\begin{algorithm}[ht!]
\small
\caption{TEMA Training}
\label{alg:TEMA-train}
\begin{algorithmic}[1]
\Require Triplets $\mathcal{T}=\{(x_r,t_m,x_t)\}_{n=1}^{N}$; frozen BLIP encoder $\Phi_{\mathbb{I}},\Phi_{\mathbb{T}}$; $\eta$; batch size $B$; hyper-parameters $\kappa,\mu$; learnable query number $N$
\Ensure Trained parameters $\Theta^\ast$
\State Initial parameters $\Theta$ 
\For{$\text{epoch}=1$ to $E$}
  \For{each mini-batch $\{(x_r^{i},t_m^{i},x_t^{i})\}_{i=1}^{B}$}
    \State \textbf{MMT Parsing Assistant (Sec~\ref{sec:pa})}$\rightarrow$
    \For{$i=1$ to $B$}
      \State Generating the summary $t_s^{i}$
      \State Check $t_s^{i}$ using Consistency Detector
    \EndFor
    \State $E_s^{i}=\Phi_{\mathbb{T}}(t_s^{i})$
    \State \textbf{MMT-oriented Entity Mapping (Sec~\ref{sec:em})}$\rightarrow$
    \State \textbf{Feature Extraction (Sec~\ref{sec:em})}$\rightarrow$: 
    \State $E_{r,g}^{i},E_{r,l}^{i}=\Phi_{\mathbb{I}}(x_r^{i})$;
    \State $E_{m,g}^{i},E_{m,l}^{i}=\Phi_{\mathbb{T}}(t_m^{i})$;
  \State$E_{t,g}^{i},E_{t,l}^{i}=\Phi_{\mathbb{I}}(x_t^{i})$
    \State \textbf{Textual Entity Mapping (Sec~\ref{sec:em})}$\rightarrow$:
    \State $\hat{\mathbf{a}}_q = \operatorname{Transformer}\left(\left[\textbf{E}_s, \textbf{E}^{l}_m, \mathbf{a}_q\right]\right)$
    \State \textbf{Visual Entity Mapping (Sec~\ref{sec:em})}$\rightarrow$:
    \State $\hat{\mathbf{b}}_q = \operatorname{Transformer}\left(\left[\textbf{E}^{g}_r, \textbf{E}^{l}_r, \mathbf{b}_q\right]\right)$
    \State \textbf{Multimodal Query Composition (Sec~\ref{sec:em})}$\rightarrow$
    \State $\hat{\textbf{E}}_m = [\textbf{E}^{g}_m, \hat{\mathbf{a}}_q]$, $\hat{\textbf{E}}_r = [\textbf{E}^{g}_r, \hat{\mathbf{b}}_q]$
    \State Then we get composed feature $\textbf{E}_c$:
    \State $\textbf{E}_c = \textbf{Combiner} (\hat{\textbf{E}}_m, \hat{\textbf{E}}_r)$
    \State \textbf{Summary-guided Distillation (Eqn~\ref{eq:summ})}$\rightarrow$
    \State $\mathcal{L}_{summ} = \operatorname{1} - \operatorname{cos}\left( \textbf{E}_s, \bar{\mathbf{a}}_q\right),
    \label{eq:cos}$
    \State \textbf{Orthogonal Regularization (Eqn~\ref{eq:ortho})}$\rightarrow$
    \State $\mathcal{L}_{ortho} =\left\| {\hat{\mathbf{a}}_q}^{\top}\hat{\mathbf{a}}_q - \mathbf{I} \right\|_{F}^{2} + 
    \left\| {\hat{\mathbf{b}}_q}^{\top}\hat{\mathbf{b}}_q - \mathbf{I} \right\|_{F}^{2}, 
    \label{eq:ortho}$
    \State \textbf{Batch-based Classification Loss (Eqn~\ref{bbc})}$\rightarrow$
    \State $\mathcal{L}_{bbc}
    =\frac{1}{B} \sum_{i=1}^{B} -\log \{ \frac{\exp \{ \operatorname{s}( \bar{\textbf{E}}_{ci}, \bar{\textbf{E}}_{ti})  / \tau\}}{ \sum_{j=1}^{B} \exp \{ \operatorname{s}( \bar{\textbf{E}}_{ci}, \bar{\textbf{E}}_{tj}) / \tau \}  } \}$
    \State \textbf{Overall Object (Eqn~\ref{eq:optimization})}$\rightarrow$
    \State $\mathcal{L}=\mathcal{L}_{\text{bbc}} + \mu\,\mathcal{L}_{\text{ortho}}$
    \State $\Theta \leftarrow \operatorname{OptimizerUpdate}(\Theta,\nabla_{\Theta}\mathcal{L})$
  \EndFor
\EndFor
\State \Return $\Theta^\ast$
\end{algorithmic}
\end{algorithm}

\begin{algorithm}[ht!]
\small
\caption{TEMA Inference}
\label{alg:TEMA-infer}
\begin{algorithmic}[1]
\Require Queries $(x_r,t_m)$; candidate images $\{x\}$; frozen BLIP encoder $\Phi_{\mathbb{I}},\Phi_{\mathbb{T}}$;
\Ensure Ranked retrieval results
\State Feature Extraction: \State $E_{r,g},E_{r,l}=\Phi_{\mathbb{I}}(x_r),\; E_{m,g},E_{m,l}=\Phi_{\mathbb{T}}(t_m)$
\State Textual \& Visual Entity Mapping: 
\State $\hat{\mathbf{a}}_q = \operatorname{Transformer}\left(\left[\textbf{E}_s, \textbf{E}^{l}_m, \mathbf{a}_q\right]\right)$
\State $\hat{\mathbf{b}}_q = \operatorname{Transformer}\left(\left[\textbf{E}^{g}_r, \textbf{E}^{l}_r, \mathbf{b}_q\right]\right)$
\State Composed Feature:
\State $\textbf{E}_c = \textbf{Combiner} (\hat{\textbf{E}}_m, \hat{\textbf{E}}_r)$
\State  For each candidate image $x$: \State $\textbf{E}_t(x)=\Phi_{\mathbb{I}}(x)$
\State Obtain the similarity:
\State $s(x)=\mathrm{sim}(\textbf{E}_c,\textbf{E}_t(x))$
\State Rank by $s(x)$
\end{algorithmic}
\end{algorithm}

\section{Algorithm of TEMA's Training and Inference Process}
\label{sup:algorithm}
To complement the method section in the full paper and more clearly illustrate the TEMA processing flow, we provide the complete TEMA training and inference processes in the form of pseudocode, which are presented in Algorithm~\ref{alg:TEMA-train} and Algorithm~\ref{alg:TEMA-infer}, respectively.

\section{More Analysis on Prompts}
\label{sup:prompt}
In this section, we provided a more detailed analysis of prompts. In Appendix~\ref{sup:detailed_prompts}, we present the detailed prompts used for MMT generation in the M-FashionIQ and M-CIRR datasets. In Appendix~\ref{sup:analysis_different_prompts}, we analyzed the effect of different prompts on MMT generation.

\begin{figure*}[t!]
  \centering
  \includegraphics[width=\linewidth]{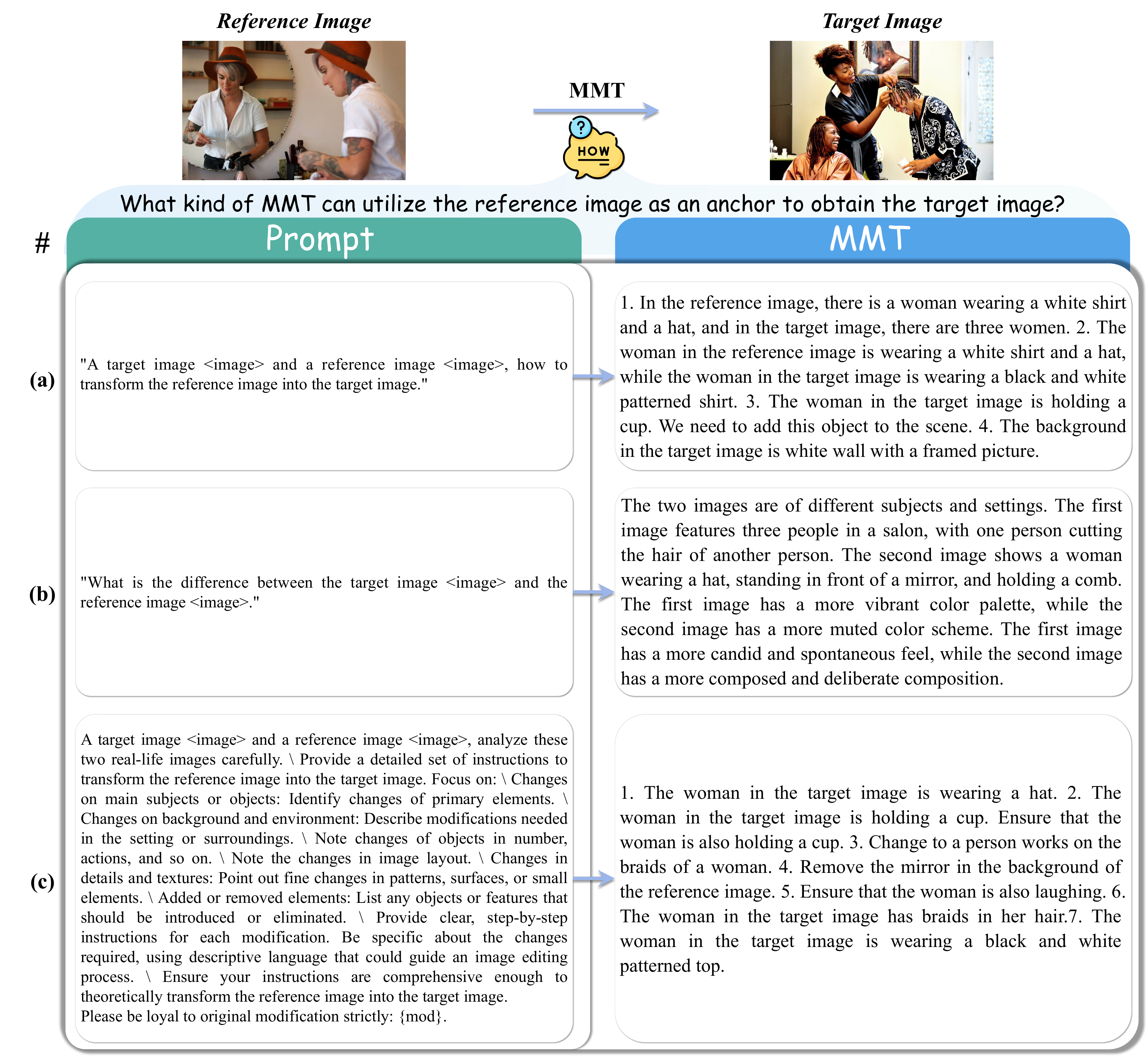}
  \caption{Generated MMTs using various prompts for BLIP-3.}
  \label{fig:diffprompt}
\end{figure*}

\begin{figure*}[t!]
  \centering
  \includegraphics[width=\linewidth]{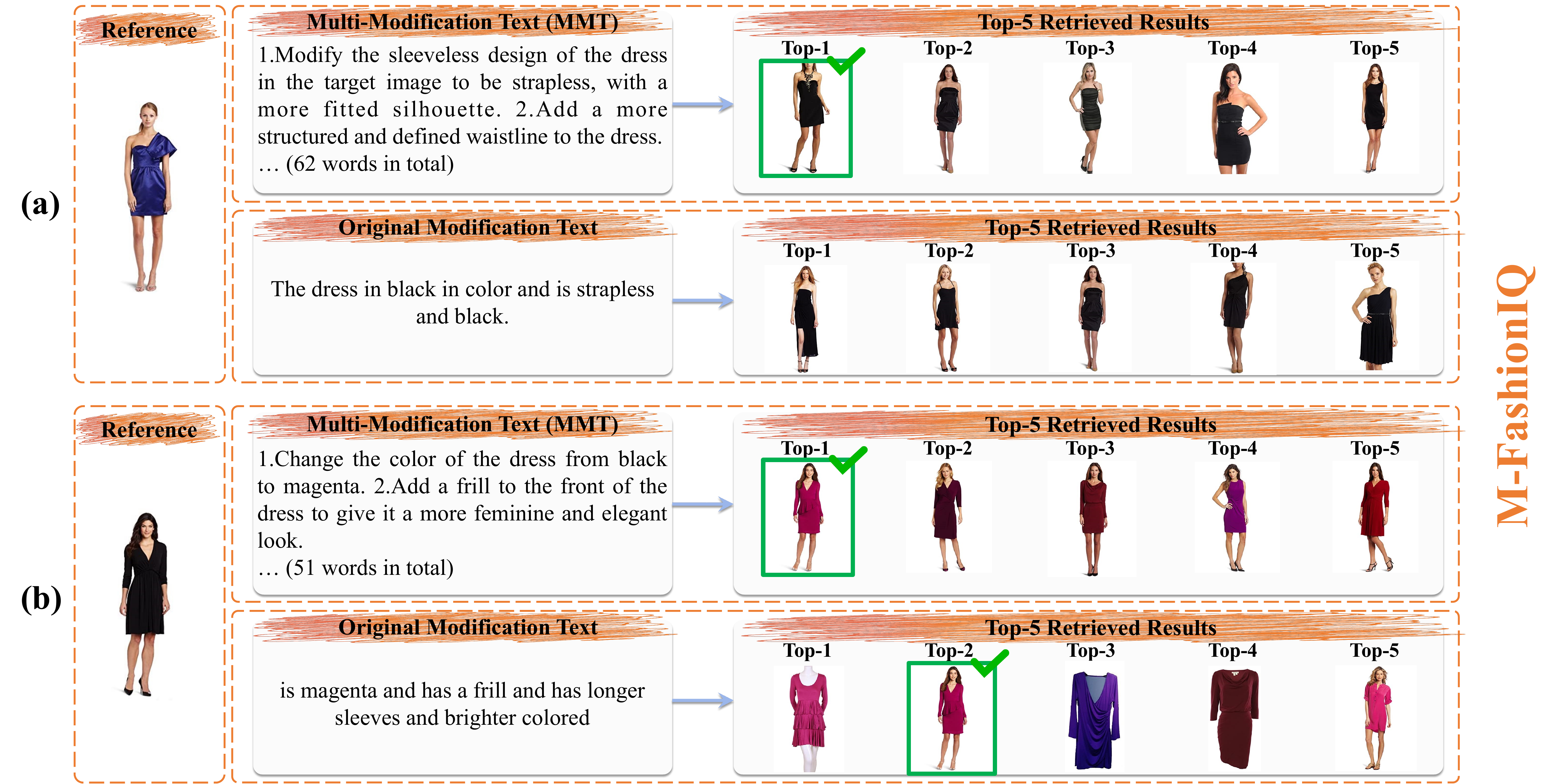}
  \caption{The mitigation on the false-negative samples when using MMT. We showed the top-5 retrieved results on both multi-modification datasets and original CIR datasets. The target images are framed in green.}
  \label{fig:supple_false_fiq}
\end{figure*}

\subsection{Detailed Prompts for MMT Generation}
\label{sup:detailed_prompts}

In Figure~\ref{fig:sup_prompt}, we showed the prompts used for generating the MMT of both M-FashionIQ and M-CIRR datasets, utilizing BLIP-3~\cite{blip3}.

For M-FashionIQ, we employed specifically designed prompts for BLIP-3. Based on the characteristics of the original FashionIQ dataset (which primarily focuses on various garment details such as the presence/absence of straps and sleeve lengths), these prompts were crafted to direct BLIP-3's output toward attention on different components of garments, holistic perception of the clothing items, and comparative analysis between reference and target images. This process enabled us to generate fine-grained MMT that captures detailed modification descriptions.

For M-CIRR, the details are more complex because the original CIRR dataset is open-domain, which generally involves several different objects as well as background elements. This also confirms the necessity of the MMT, which is only detailed enough to clearly describe the real modifications based on the reference and target images. Specifically, we require that the output of BLIP-3 focuses on changes in main subjects or objects, backgrounds and environments, details and textures, and so on~\cite{ma2024context,long2026aisupervisorautonomousairesearch,xu2025reducing,song4,fu2026maspo}.

\begin{figure*}[t!]
  \centering
  \includegraphics[width=\linewidth]{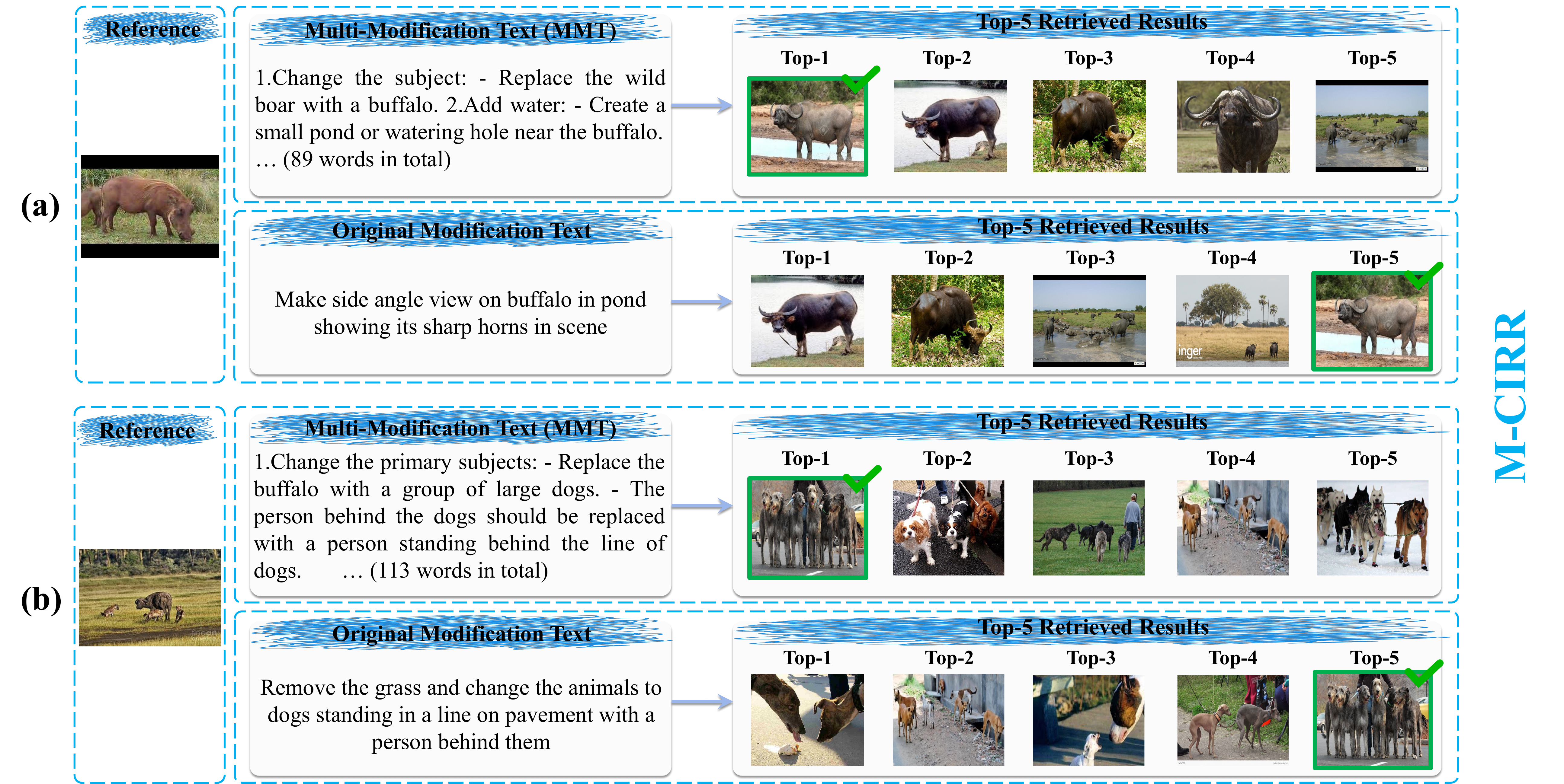}

  \caption{The mitigation on the false-negative samples when using MMT. We showed the top-5 retrieved results on both multi-modification datasets and original CIR datasets. The target images are framed in green.}

  \label{fig:supple_false_cirr}
\end{figure*}

\subsection{Analysis of Different Prompts}
\label{sup:analysis_different_prompts}
Additionally, we require that the output of BLIP-3 be faithful to the original short modification text and that the MMT expands upon it, including details not noted in the original modification text. For example, the original modification text may refer to one object in the whole scene, however, the target image actually changes more than one object based on the reference image, allowing the MMT to describe the complete and detailed modification, avoiding the incompleteness of the original modification text.

For a more detailed analysis, we report different MMTs using various designed prompts for BLIP-3, as shown in Figure~\ref{fig:diffprompt}. In this figure, for prompts (a) and (c), both of them well captured the detailed modification from the reference image to the target image and elaborated on different perspectives. 
However, the MMT based on prompt (c) is more precise and comprehensive, focusing on one-to-many mappings and multiple constraints. Thus, we choose prompt (c) for our pipeline.

\section{More Qualitive Results}
\label{sup:more_qualitive_results}
\subsection{Mitigation on False-negative Samples}
\label{sup:false_negative}
Ground-truth labeling in the CIR datasets are often insufficient due to the presence of numerous visually similar images and the limited descriptive capability of short modification text. For a given multimodal query, there are candidate images that differ subtly from the ground-truth yet still meet the query, these images are all labeled as negative samples, which we refer to as false-negative samples.

To validate the advantages of our proposed M-FashionIQ and M-CIRR in reducing false-negative samples, we selected a straightforward baseline, BLIP4CIR~\cite{blip4cir}, which incorporates multimodal query features to derive compositional features for retrieving target images. We conducted experiments in the MMT scenario for multi-modification datasets, and the original modification text scenario for original CIR datasets. As illustrated in Figure~\ref{fig:supple_false_fiq} and Figure~\ref{fig:supple_false_cirr}, we present the top-5 retrieval results for both scenarios, with the target images highlighted in green boxes.

In the fashion domain dataset M-FashionIQ, as illustrated by the CIR example (bottom) in Figure~\ref{fig:supple_false_fiq}(a), the top-5 results retrieved by the model satisfy the multimodal query, however, they are all classified as negative samples. This occurs because the short modification text in CIR fails to account for the inclusion of ``a necklace'' in the target image. Conversely, the multi-modification annotation example (top) in Figure~\ref{fig:supple_false_fiq}(a) highlights this detail modification after employing MMT relabeling, resulting in the target image becoming the only positive sample and significantly reducing the number of false-negative samples. A similar situation is observed in Figure~\ref{fig:supple_false_fiq}(b), where the ranking of the target image in the retrieval results improves from second to first position following relabeling by MMT.

For open domain dataset M-CIRR, as the case shown in Figure~\ref{fig:supple_false_cirr}(a), the original short modification text only required for ``side angle view on buffalo'', ``in pond'', and ``sharp horns''. However, the ground-truth (retrieved successfully using the MMT) included more requirements, such as ``standing in the water'', ``dirt path'', and ``trees in the background''. In the multi-modification scenario, the MMT encapsulated these details that are not present in the original short modification text, and therefore correctly retrieved the target image, weakening the impact caused by the false negative issue. Similarly, in Figure~\ref{fig:supple_false_cirr}(b), the ranking of the target image in the retrieval results improves from fifth to first position after relabeling by MMT. These results demonstrate that MMT provides more detailed descriptions, causing the original false-negative samples to no longer satisfy the new multimodal query. Thus, these samples are converted into true-negative samples, alleviating the issue of false negatives in the multi-modification datasets and reducing their impact on model training.

\begin{figure*}[t!]
  \centering
  \includegraphics[width=\linewidth]{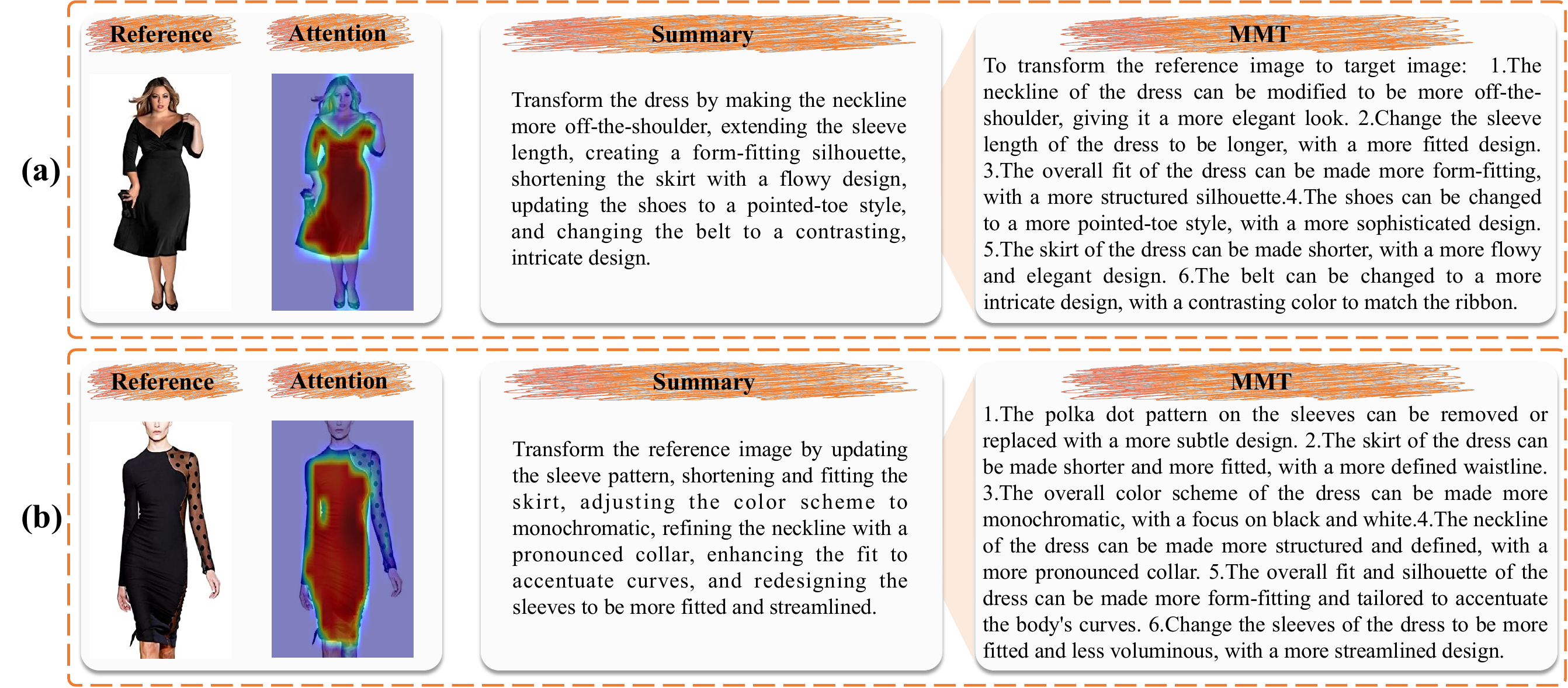}

  \caption{Attention visualization results for the reference image on M-FashionIQ by the PA-generated summary.}

  \label{fig:pa_attn_fiq}
\end{figure*}

\begin{figure*}[t!]
  \centering
  \includegraphics[width=\linewidth]{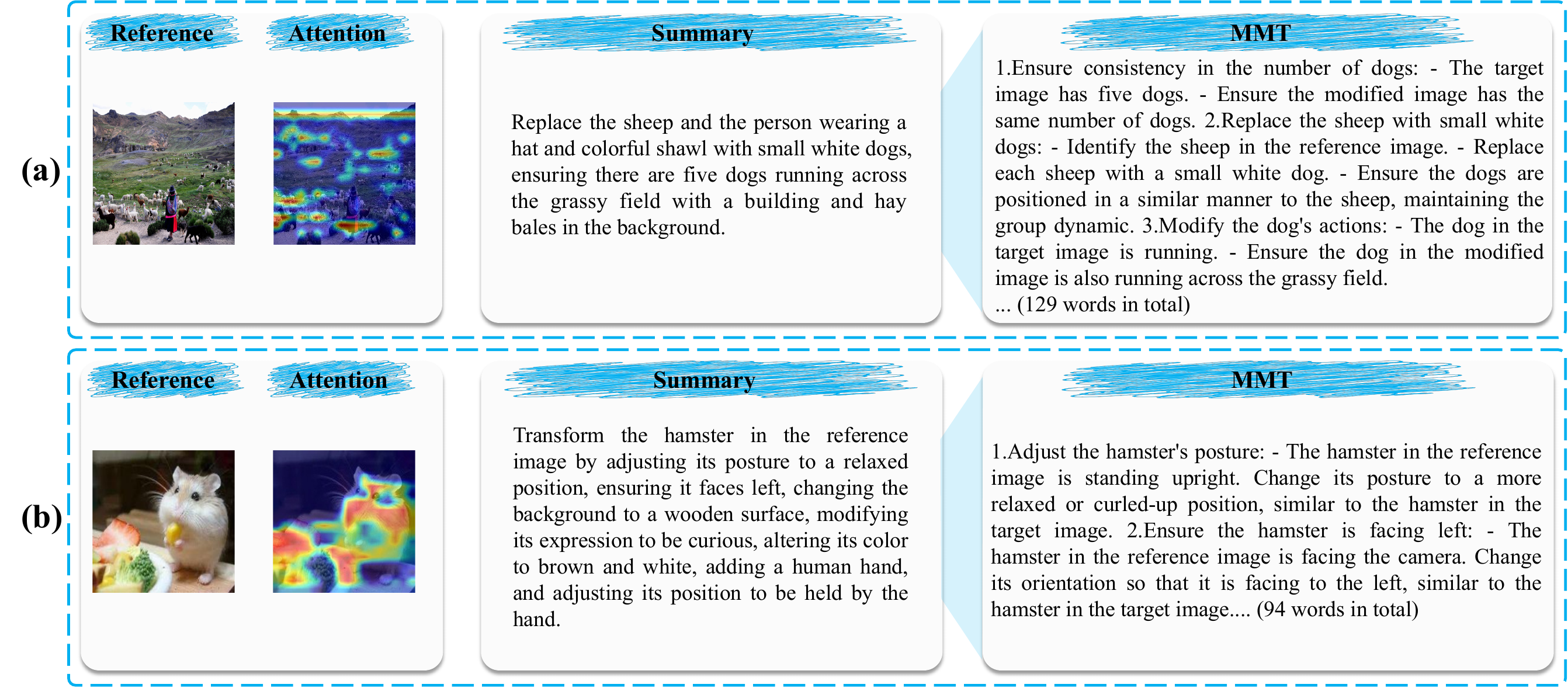}

  \caption{Attention visualization results for the reference image on M-CIRR by the PA-generated summary.}

  \label{fig:pa_attn_cirr}
\end{figure*}

\subsection{Attention Visualization for Summaries}
\label{sup:summary_vis}
The summary generated by the PA module serves as a simplified representation in MMT, encompassing all the to-be-modified entities in the reference image. To evaluate the quantity of the summaries, we employed Grad-CAM to visualize its attention in relation to the reference image.

For the fashion domain dataset M-FashionIQ, as the case illustrated in Figure~\ref{fig:pa_attn_fiq}(a), the to-be-modified entities in MMT include ``neckline'', ``sleeve length'', ``shoes'', ``skirt'', and ``belt''. The summary concisely consolidates these entities into a single sentence while omitting specific modification details for each entity. The attention visualization demonstrates the summary's focus on different regions of the reference image. We observed that all to-be-modified entities are well-attended, validating the correctness and accuracy of the summary content. Similarly, the summary in Figure~\ref{fig:pa_attn_fiq}(b) highlights all the to-be-modified entities, including ``sleeve pattern'', ``skirt'', and ``neckline''.

For the open domain dataset M-CIRR, taking the case in Figure~\ref{fig:pa_attn_cirr}(a) as an example, 
the MMT includes the to-be-modified entities ``sheep'' and ``person'', with many modification details. In contrast, the summary succinctly and accurately expresses the to-be-modified entities while omitting detailed information. 
Through the attention visualization, we observe that both the ``sheep'' distributed across different regions and the single ``person'' are well-attended to. Similarly, the summary in Figure~\ref{fig:pa_attn_cirr}(b) addresses the full range of to-be-modified entities in MMT, including the ``hamster'', ``background'', and ``expression''.
This validated the effectiveness of our generated summary in encompassing all the to-be-modified entities in the MMT, thereby enhancing the model's focus on these entities.                                                             

\begin{figure*}[h]
  \centering
  \includegraphics[width=\linewidth]{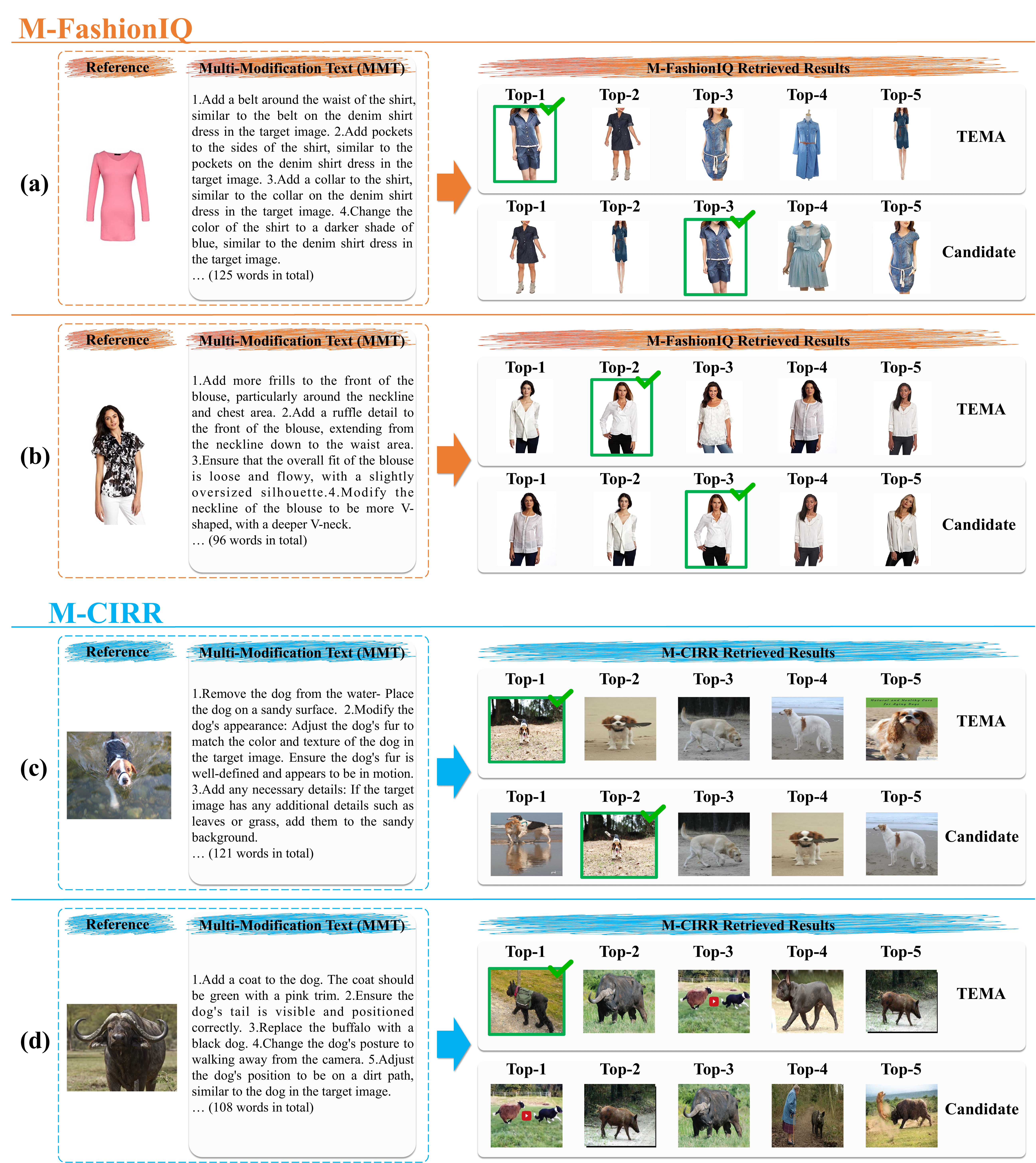}
  \caption{Qualitative examples of our proposed TEMA compared to the sub-optimal model Candidate.}

  \label{fig:case}
\end{figure*}

\subsection{Case Study}
\label{sup:quali}
To intuitively validate the performance of our proposed TEMA on multi-modification datasets, we present several examples demonstrating TEMA's retrieval results, along with the comparison to sub-optimal model candidates, as shown in Figure~\ref{fig:case}. 
Specifically, Figure~\ref{fig:case}(a) and (b) demonstrated the results on M-FashionIQ, while Figure~\ref{fig:case}(c) and (d) present the results on M-CIRR. 

For cases in Figure~\ref{fig:case}(a), (c), and (d), TEMA successfully retrieved the target images at top-1, whereas Candidate failed to rank the target images in the first position. For the case in Figure~\ref{fig:case}(d), the Candidate model even failed to retrieve the target image within the top-5 results. These examples demonstrate the superior performance of our proposed TEMA and its effectiveness. Notably, in Figure~\ref{fig:case}(b), both TEMA and Candidate failed to retrieve the target image in the first position, which may be attributed to insufficient annotation in the original CIR dataset. While our proposed TEMA framework mitigated the false-negative issues inherent in CIR, these problems persisted in a small number of triplets, resulting in model failure in these cases.